\documentclass[10pt,twocolumn,letterpaper]{article}

\usepackage{cvpr}              % To produce the CAMERA-READY version
\usepackage{graphicx}
\usepackage{amsmath}
\usepackage{amssymb}
\usepackage{booktabs}
\usepackage{multirow}
\usepackage{float}

\usepackage[pagebackref,breaklinks,colorlinks]{hyperref}

\usepackage[capitalize]{cleveref}
\crefname{section}{Sec.}{Secs.}
\Crefname{section}{Section}{Sections}
\Crefname{table}{Table}{Tables}
\crefname{table}{Tab.}{Tabs.}

\def\confName{CVPR}
\def\confYear{2022}

\begin{document}

% \title{Quadratic Search Network for Highly-Efficient Multi-View Stereo}
\title{Generalized Binary Search Network for Highly-Efficient Multi-View Stereo}

\author{Zhenxing Mi \and Di Chang \and Dan Xu \and
The Department of Computer Science and Engineering, HKUST\\
{\tt\small \{zmiaa, dchangac\}@connect.ust.hk,}
{\tt\small danxu@cse.ust.hk}
}
\maketitle

%%%%%%%%% ABSTRACT
\begin{abstract}
Multi-view Stereo (MVS) with known camera parameters is essentially a 1D search problem within a valid depth range.~Recent deep learning-based MVS methods typically densely sample depth hypotheses in the depth range, and then construct prohibitively memory-consuming 3D cost volumes for depth prediction. Although coarse-to-fine sampling strategies alleviate this overhead issue to a certain extent, the efficiency of MVS is still an open challenge. In this work, we propose a novel method for highly efficient MVS that remarkably decreases the memory footprint, meanwhile clearly advancing state-of-the-art depth prediction performance.~We investigate what a search strategy can be reasonably optimal for MVS taking into account of both efficiency and effectiveness. We first formulate MVS as a binary search problem, and accordingly propose a generalized binary search network for MVS. Specifically, in each step, the depth range is split into 2 bins with extra 1 error tolerance bin on both sides.~A classification is performed to identify which bin contains the true depth. We also design three mechanisms to respectively handle classification errors, deal with out-of-range samples and decrease the training memory. The new formulation makes our method only sample a very small number of depth hypotheses in each step, which is highly memory efficient, and also greatly facilitates quick training convergence. Experiments on competitive benchmarks show that our method achieves state-of-the-art accuracy with much less memory. Particularly, our method obtains an overall score of 0.289 on DTU dataset and tops the first place on challenging Tanks and Temples advanced dataset among all the learning-based methods. 
The trained models and code will be released at \url{https://github.com/MiZhenxing/GBi-Net}.
\end{abstract}

%%%%%%%%% BODY TEXT
\section{Introduction}
\label{sec:intro}
Multi-view Stereo (MVS) is a long-standing and fundamental topic in computer vision, which aims to reconstruct 3D geometry of a scene from a set of overlapping images~\cite{furukawa2009accurate, vu2011high, galliani2015massively, schonberger2016pixelwise, xu2019multi}.
With known camera parameters,
MVS matches pixels across images to compute dense correspondences and recover 3D points, which is essentially a 1D search problem~\cite{10.1561/0600000052}.
A depth map is widely used as 
3D representation due to its regular format.~To overcome the issue of coarse matching in previous purely geometry-based methods, recent learning-based MVS methods~\cite{im2019dpsnet, luo2019p, yao2018mvsnet, yao2019recurrent} designed deep networks for dense depth prediction to significantly advance traditional pipelines.
\begin{figure}[t]
  \centering
  \includegraphics[width=\linewidth]{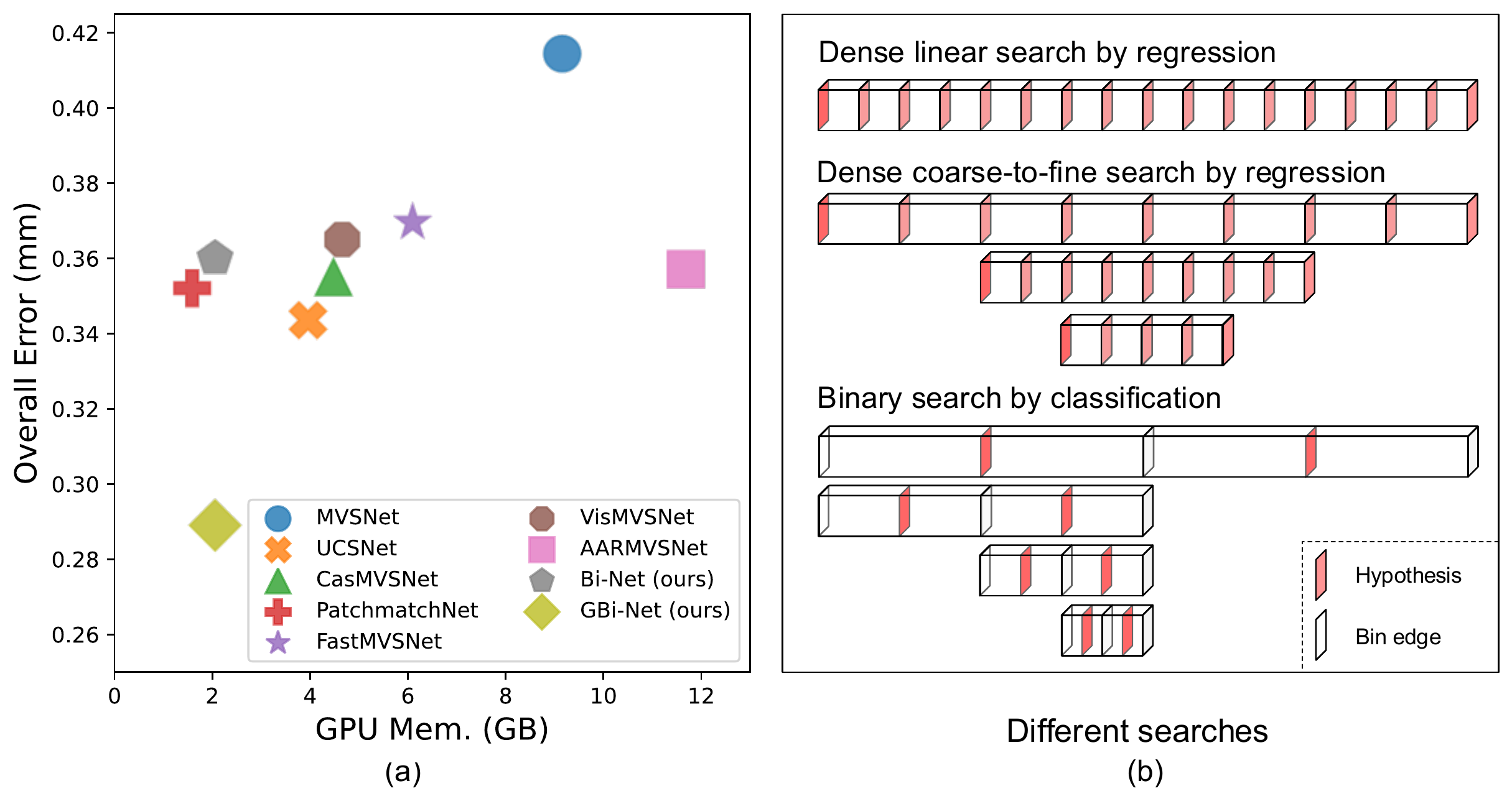}
  \vspace{-20pt}
  \caption{(a) Comparison with previous state-of-the-art learning-based
MVS methods~\cite{yao2018mvsnet,wang2021patchmatchnet,Cheng_2020_CVPR,gu2020cascade,Yu_2020_CVPR,zhang2020visibilityaware,Wei_2021_ICCV} on DTU~\cite{aanaes2016large}.
The relationship between the overall error and the GPU memory overhead with image size $1152 \times 1600$ and image number $5$. (b) Comparison of the previous dense search and the proposed binary search.
}
  \label{fig:motivation}
  \vspace{-15pt}
\end{figure}
% \begin{figure}[t]
%      \centering
%   \begin{minipage}[htbp]{0.5\textwidth}
%      \centering
%      \includegraphics[width=0.4\linewidth]{image/memory.pdf}
%      \caption{Interpolation for Data 1}\label{Fig:Data1}
%   \end{minipage}
% %   \hfill
%   \begin{minipage}[htbp]{0.5\textwidth}
%      \centering
%      \includegraphics[width=0.4\linewidth]{image/differentsearchs.pdf}
%      \caption{Interpolation for Data 2}\label{Fig:Data2}
%   \end{minipage}
% \end{figure}
For instance, MVSNet~\cite{yao2018mvsnet} and RMVSNet~\cite{yao2019recurrent} propose to construct 3D cost volumes 
from 2D image features with dense depth hypotheses.
A 3D cost volume is a 5D tensor and is typically regularized by a 3D Convolutional Neural Network (CNN) or a Recurrent Neural Network (RNN) for depth prediction.
% has a spatial size of $D$, $H$
% and $W$, where $D$ is the depth number and $H$, $W$
% is the resolution of 2D feature maps.
% The 3D cost volumes are then regularized 
% by 3D CNN or recurrent neural networks
% for depth prediction.
\par The importance of 3D cost volume regularization for accurate depth prediction has been confirmed by other works \cite{chen2019point, Cheng_2020_CVPR, gu2020cascade}.
% reveal the importance of 3D cost volume regularizations
% for accurate predication.
However, a severe problem is that 3D cost volumes
are highly memory-consuming.
Existing works made significant efforts to address this issue via decreasing the resolution of feature maps \cite{yao2018mvsnet}, using a coarse-to-fine strategy that gradually 
increases resolution of feature maps while
decreasing the depth hypothesis number \cite{chen2019point, Cheng_2020_CVPR, gu2020cascade}, and removing expensive 3D CNN or RNN \cite{wang2021patchmatchnet, yang2021mvs2d}. Although the memory can be alleviated to some extent, relatively lower accuracy is commonly observed. The size of 3D cost volume, specifically the depth hypothesis number, plays a dominant role in causing a large memory footprint.
% One straightforward solution is to decrease the spatial resolution of the feature maps, which yet inevitably
% decreases the depth accuracy~\cite{yao2018mvsnet}. 
% Some other methods ~\cite{chen2019point, Cheng_2020_CVPR, gu2020cascade}
% use a coarse-to-fine pipeline which gradually 
% increases the resolution of feature maps while
% decreasing the depth channels, and some other attempt to remove 3D CNNs or RNNs~\cite{wang2021patchmatchnet, yang2021mvs2d} 

% which can alleviate the memory issue to some extent. The attempts of removing 3D CNNs or RNNs~\cite{wang2021patchmatchnet, yang2021mvs2d} can obtain better efficiency 
% could get higher efficiency while obtaining relatively
% lower accuracy.
% From these methods, we could see that
% the size of 3D cost volume, more specifically the $D$,  
% is a dominant factor in memory consumption.

% \begin{figure}[t]
%   \centering
%   \includegraphics[width=\linewidth]{image/differentsearchs.pdf}

%   \caption{Illustration of different searches.}
%   \label{fig:differentsearchs}
% \end{figure}
Due to the significance of 3D cost volumes in both model efficiency and effectiveness, a critical question naturally arises: what is a minimum volume size to secure satisfactory accuracy while maintaining as small as possible the memory overhead?
% Driven by the significance of 3D cost volumes, a question arises: what is the minimum size of the volume 
% do we need for good reconstruction results?
% This question can be seen as how to find the optimal number of depth hypotheses. 
In this work, we investigate this question by exploring from a perspective of discrete search strategies, to identify a minimal depth hypotheses number, a key factor in 3D cost volumes. 
% We could analyse this problem from a perspective of search algorithm. 
As shown in Fig.~\ref{fig:motivation}b,
the vanilla MVSNet~\cite{yao2018mvsnet} can be seen as a dense search method that checks all depth hypotheses similar to a linear search in a parallel manner. The coarse-to-fine methods~\cite{gu2020cascade, Cheng_2020_CVPR} perform a multi-granularity search, which starts from a coarse level
and gradually refines the prediction. However, these two types of methods both consider dense search in each stage. We argue that the dense search does not necessarily guarantee better accuracy due to a much larger prediction space and significantly increases model complexity, leading to higher optimization difficulty in model training. 

\par To explore a reasonably optimal search strategy, we first formulate MVS as a binary search problem, which can remarkably reduce the cost volume size to an extremely low bound. It performs comparisons and eliminates half of the search space in each stage (see Fig.~\ref{fig:motivation}b), and can convergence quickly to a fine granularity within logarithmic stages. 
% is quite efficient to find a target
% value in a sorted list, 
% while depth hypotheses in MVS are naturally ordered
% and sorted.
% As shown in Figure \ref{fig:differentsearchs},
% ideally the binary search only performs 
% 1 comparison and eliminate half of the search space.
% in each iteration to determine which bin the target value is in.
% It convergences quickly to fine granularity in 
% logarithmic iterations. 
% The idea of binary search shows promising property to
% largely decrease the size of 3D cost volumes if it is
% successfully used in MVS.
% maximum 2 comparisons
% In this paper we propose to
% leverage binary search in MVS.
% The basic idea is to perform
% ``comparisons" by networks.
% With the comparison results, binary search
% is used for updating
% prediction results till the predictions are fine enough.
In contrast to regression-based methods, which directly sample depth values from the depth range, we first divide the depth range into 2 bins.
In our design, the `comparisons' by the network is to determine which bin contains the true depth value via performing a \emph{binary} discrete classification. We use \emph{center} points of the two bins to represent them and construct 3D cost volumes. 
The binary search offers superior efficiency, while it also brings an issue of the network classification error (\ie~prediction out of bins). The error can be accumulated from each search stage causing unstable optimization gradients and relatively low accuracy.

\par To tackle this issue, we further design three effective mechanisms, and accordingly propose a generalized binary search deep network, termed as GBi-Net, for highly efficient MVS. 
The first mechanism is that we pad two error tolerance bins on the two sides to reduce the prediction error of out of bins.
The second mechanism is for training. If the network generates an error of prediction out of bins for a pixel at a search stage, we stop the forward pass of this pixel in the next stage, and the gradients at this stage are not used to update the network. Extensive experiments show that the proposed GBi-Net can 
largely decrease the size of 3D cost volumes for a significantly efficient network, and more importantly, without any trade-off on the depth prediction performance. The third is efficient gradient updating. It updates the network parameters immediately at each search stage without accumulating across different stages as most works do. It can largely reduce the training memory while maintaining the performance.
% we could successfully train the 
% search network and largely
% improve the reconstruction results.
% With this network, we decrease the
% size $D$ of 3D cost volumes to $4$, which largely decreases the memory
% usage in learning-based MVS without trade-off on accuracy.
Our method achieves state-of-the-art performance on different competitive datasets including DTU~\cite{aanaes2016large} and Tanks \& Temples~\cite{knapitsch2017tanks}. Notably, 
% show that our network could reconstruct accuracy 3D
% geometry for different dasatsets.
on DTU, we achieve an overall score of 
\textbf{0.289} (lower is better), 
remarkably improving the previous best performing method, 
and also obtain a memory efficiency improvement 
of \textbf{48.0\%} compared to UCSNet \cite{Cheng_2020_CVPR}
and \textbf{54.1\%} 
compared to CasMVSNet \cite{gu2020cascade} (see Fig.~\ref{fig:motivation}a).

In summary, our contribution is three-fold:
\vspace{-3pt}
\begin{itemize}
\setlength{\itemsep}{0pt}
\setlength{\parsep}{0pt}
\setlength{\parskip}{3pt}
    \item We investigate efficient MVS from a perspective of search strategies, and propose a discrete binary search method for MVS (Bi-Net) which can vastly decrease the memory usage of 3D cost volumes.
    \item We design a highly-efficient generalized binary search network~(GBi-Net) via further designing three mechanisms (\ie~padding error tolerance bins, gradients masking, and efficient gradient updating) with the binary search to avoid error accumulation of false network predictions and improve efficiency. 
    \item We evaluate our method on several challenging MVS datasets
    and significantly advance existing state-of-the-art methods in terms of both the depth prediction accuracy and the memory efficiency.
\end{itemize}

\begin{figure*}[t]
  \centering
   \includegraphics[width=\linewidth]{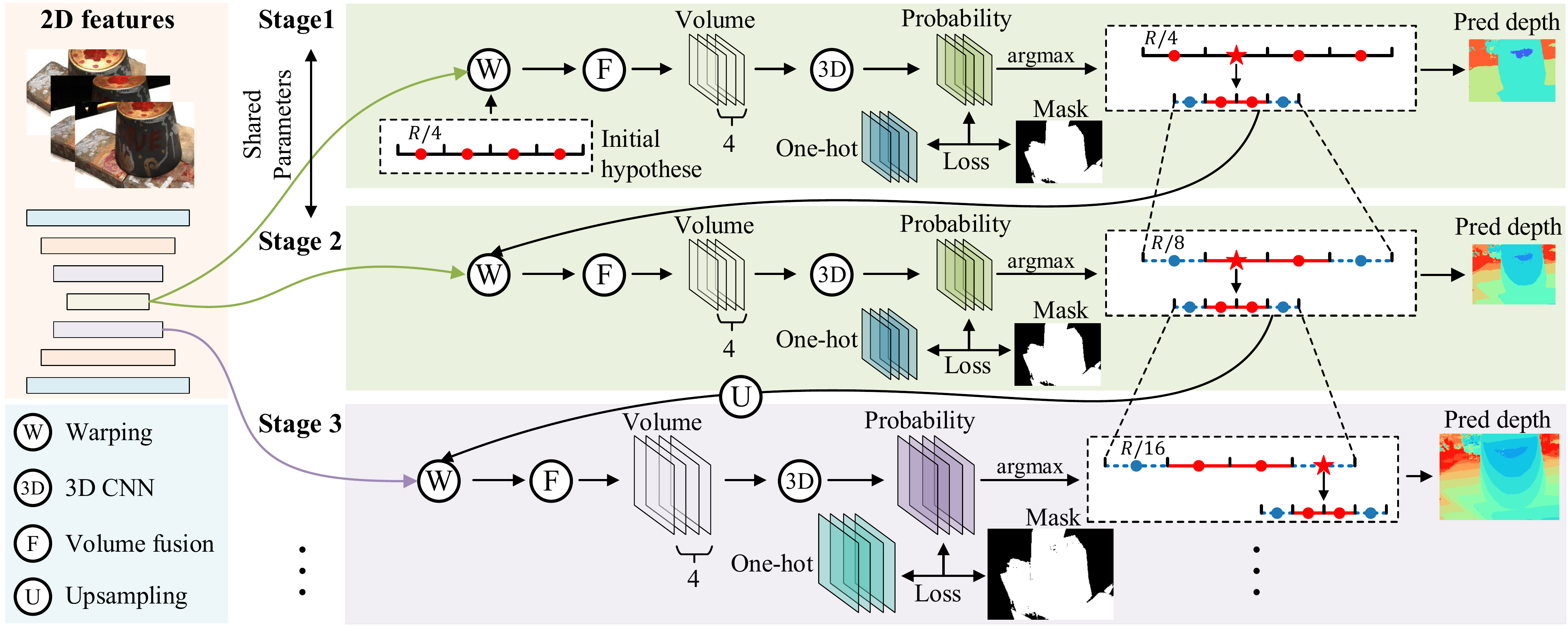}
    \vspace{-16pt}
   \caption{The multi-stage framework of our GBi-Net. The 2D CNNs first extract 2D feature maps. Then a 3D cost volume is constructed and fused by differentiable warping and only $4$ depth hypotheses. The cost volume is regularized by 3D CNNs and gets a probability volume for loss calculation with one-hot labels and training masks. We select the depth bin at current stage using $argmax$ operation. Then we update the depth hypotheses for the next stage with our proposed
   Generalized Binary Search. 
%   The detailed updating step can be seen in Fig.\ref{fig:quadraticsearch}.
}
   \vspace{-10pt}
   \label{fig:pipelinemain}
\end{figure*}

\vspace{-8pt}
\section{Related Work}
\label{sec:formatting}
We review the most related works in the literature from two aspects, \ie~traditional MVS and learning-based MVS. 
% \subsection{Traditional Multi-View-Stereo}
\par\noindent\textbf{Traditional Multi-View Stereo.} In 3D reconstruction, various 3D representations are used, such as volumetric representation~\cite{kutulakos2000theory,seitz1999photorealistic}, point cloud~\cite{furukawa2009accurate,lhuillier2005quasi}, mesh~\cite{curless1996volumetric,kazhdan2006poisson,tang2019skeleton,tang2021skeletonnet} 
and depth map~\cite{galliani2015massively,schonberger2016pixelwise,campbell2008using}. 
% Volume-based methods first discretize the entire 3D space into regular cubes, and then use the photometric consistency metric to determine whether the voxel belongs to the surface. Spatial discretization is memory tight, so these methods cannot be extended to large-scale scenarios. Point-based methods focus on 3D points, usually starting from a set of sparse matching key points, using propagation strategies to gradually dense reconstruction, which limits the ability of parallel data processing. Mesh-based methods build completely out of tris or polys. These methods create more “high-fidelity” designs (smoother looking surfaces, more details, etc.). They can be deformed and animated more naturally, also easy to render and visualize. 
In MVS\cite{merrell2007real,schonberger2016pixelwise,xu2019multi}, depth maps have shown advantages in robustness, efficiency.
% , and generalization capability in modeling 3D scenes. 
% converts the task of reconstructing the 3D scene into 
They estimate a depth map of each 
reference image and fuse them into one 3D point cloud. 
% Several successful traditional MVS algorithms utilizing depth maps have been proposed.
For instance, COLMAP~\cite{schonberger2016pixelwise} simultaneously estimates pixel-wise view selection, depth, and surface normal. ACMM~\cite{xu2019multi} leverages a multi-hypothesis joint voting scheme for view selection from different candidates. These existing methods perform modeling of occlusion, illumination across neighboring views for depth estimation. Although stable results can be achieved, high matching noise and poor correspondence localization in complex scenes are still severe limitations. Thus, our method mainly focuses on developing a deep learning-based MVS pipeline to advance the estimation performance.  
% which further inspires our work to model the 3D scene information in learning-based multi-view stereo methods.

% \subsection{Learning-based Multi-View-Stereo}
\par\noindent\textbf{Learning-based~Multi-View Stereo.} Deep learning-based MVS methods~\cite{ji2017surfacenet, Huang_2018_CVPR, yao2018mvsnet,Wei_2021_ICCV} recently have achieved remarkable performance. 
% Unlike the traditional methods which use a sparse set of matched key points to recover depth maps for each reference image based on hand-crafted features, 
Deep learning-based methods usually utilize Deep CNNs to estimate a dense depth map. 
% In \cite{Huang_2018_CVPR}, disparity maps are produced directly from a set of images with camera poses among them. After the patch-matching process, Intra-volume and Inter-volume features are aggregated by ConvNets. 
Recently, 3D cost volumes have been widely used in MVS~\cite{yao2018mvsnet, wang2021patchmatchnet,zhang2020visibilityaware,xu2020pvsnet}. As a pioneering method, MVSNet~\cite{yao2018mvsnet} constructs the 3D cost volume from feature warping and regularizes the cost volume with 3D CNNs for depth regression. 
% after warping features via differentiable homography from different views of the same 3D scene.
% It proposes a popular pipeline with 2D feature extraction, 
% 3D cost volume regularization and depth regression.
The main problem of vanilla MVSNet is the large memory consumption of the 3D cost volume.
Recurrent MVSNet architectures~\cite{yao2019recurrent, yan2020dense, Wei_2021_ICCV} 
leverage recurrent networks to regularize cost volumes,
which can decrease the memory usage to some extent in the testing phase. However, the major overhead from the 3D cost volumes is not specifically addressed by these existing methods.

% Another line of learning-based methods 
% is to increase the accuracy of MVS by estimating visibility information,
% inspired from view selection in traditional MVS methods.
% PVSNet\cite{xu2020pvsnet} presents pixel-wise visibility \
% network to learn cross-view visibility information between reference view and different source views. Similar strategies are also proposed in \cite{wang2021patchmatchnet,zhang2020visibilityaware,Wei_2021_ICCV}. 
% The key of these methods is to construct two-view cost volumes. Then pixel-wise visibility maps could be regressed from them. 
% The multiple cost volumes are then fused with the visibility maps as weights. 

\par To reconstruct hig-resolution depth maps meanwhile obtaining a memory-efficient cost volume, cascade-based pipelines are proposed~\cite{gu2020cascade, Cheng_2020_CVPR,yang2020cost}, considering a coarse-to-fine \emph{dense} search strategy to gradually refine the depths. For instance, CasMVSNet~\cite{gu2020cascade} utilizes coarse feature maps and depth hypotheses in the first stage for coarse depth prediction, and then upsamples depth maps and narrows the depth range for fine-grained prediction in the next stage. 
% UCSNet~\cite{Cheng_2020_CVPR} introduces similar idea, but proposes adaptive construction by variance-based uncertainty estimation. The recurrent and cascade methods use less memory to some extent. However, the memory-consumption of 3D cost volume is still a problem in the situation of limited resources.
% ~\cite{wang2021patchmatchnet} borrows from the traditional PatchMatch methods~\cite{barnes2009patchmatch} to 
Patchmatchnet ~\cite{wang2021patchmatchnet} learns adaptive propagation and evaluation for depth hypotheses. 
It removes heavy regularization of 3D cost volumes to achieve an efficient model while 
it makes a significant trade-off between efficiency and accuracy. 

\par However, these existing works still consider a dense search in each regression stage. The memory overhead on the expensive 3D cost volume is clearly not optimized, while the proposed method targets highly efficient MVS with the designed binary search network, 
which largely advances the model efficiency, and more importantly, 
without sacrificing any depth prediction performance.    

\section{The Proposed Approach}
\label{sec:method}
In this section, we introduce the detailed structure of the proposed Generalized Binary Search Network (GBi-Net) for highly-efficient MVS. The overall framework is depicted in Fig.~\ref{fig:pipelinemain}.
% shows the iterative framework of our method. 
It mainly consists of two parts, \ie~a 2D CNN network for learning visual image representations, and the generalized binary search network for iterative depth estimation. 
% The Quadratic Search Network
% adopts a iterative search framework to
% reconstruct a accurate depth map for the reference
% image.
The GBi-Net contains $K$ search stages. In each search stage, we first compute 3D cost volumes by differentiable warping between 
the reference and source feature map in a specific corresponding scale.
Then 3D cost volumes are regularized by 3D CNNs for depth label
prediction. 
The Generalized Binary Search is responsible for initializing and updating depth hypotheses according to the predicted labels iteratively. 
% We adopt the group-wise correlation 
% \cite{guo2019group} and pixel-wise weight 
% \cite{xu2020pvsnet, wang2021patchmatchnet}
% in cost volume construction. 
% 3D Convolutional Neural Networks 
% (3D CNNs) are applied to cost volumes for regularization
% and label prediction.
% It uses predicted
% labels to update the depth hypotheses for the next iteration.
In every two stages, the networks deal with the same scale of feature maps, and the network parameters are shared. 
% If the scale in the next iteration is larger than
% the previous one, we directly upsample the predicted
% labels and then generate depth hypotheses.
Finally, one-hot labels for training the whole network are computed from ground-truth depth maps. 
% A standard cross-entropy loss is applied between the probability and the 
% one-hot labels. 
% Our proposed MVSNet network is embedded into a
% Quadratic Search framework. The networks in
% the framework is responsible for label
% prediction. 
In the next, we first introduce the 2D image encoder in Sec.~\ref{sec:imageencoding} and 
the 3D cost volume regularization 
in Sec.~\ref{sec:costvolumeregularization}.
Then, we elaborate on details about our proposed Binary Search for MVS and Generalized Binary Search
for MVS in Sec.~\ref{sec:binarysearch}
and Sec.~\ref{sec:generalizedbinarysearch}, respectively. Finally, we present the overall network optimization in Sec.~\ref{sec:networkoptimization}.
% and depth map fusion in Sec.~\ref{sec:depthmapfusion}.

% give the details of each step in one iteration.
% Our operations for each pixel are all the same,
% we do not use sub-indices for pixels in the notations,
% except the equations in Section \ref{sec:costvolumeregularization}.

\subsection{Image Encoding}\label{sec:imageencoding}
The input consists of $N$ images $\{\mathbf{I}_i\}_{i=0}^{N-1}$. $\mathbf{I}_0$ is a reference image and $\{\mathbf{I}_i\}_{i=1}^{N-1}$ is a set of $N-1$ source 
images. We use Feature Pyramid Network (FPN)~\cite{lin2017feature} as an image encoder to learn generic representation for the images with shared network parameters. From FPN, we obtain a pyramid of feature maps with 4 different scales. To have more powerful representations of the images, one deformable convolutional network (DCN)~\cite{dai2017deformable} layer is used as output layer for each scale to more effectively capture scene contexts that are very beneficial for the MVS task.
% finally producing a set of four multi-scale feature maps. 
% To change the scale of the feature maps, 3 convolution layers with stride 2 is utilized. 
% We use 3 convolution 
% layers with stride 2
% in order to change the scale of the 
% feature maps. 
% Finally, we obtain a pyramid of 
% feature maps with 4 different scales.
% For each scale, a deformable convolutional network (DCN)~\cite{dai2017deformable} is employed as an
% output network layer to learn more representative image features.
\subsection{Cost Volume Regularization} \label{sec:costvolumeregularization} 
The construction of 3D cost volumes is a critical step for deep learning-based MVS~\cite{yao2018mvsnet}. We present details about the cost volume construction and regularization for the proposed generalized binary search network. Given $D$ depth hypotheses at the $k$-th search stage, \ie~$\{d_{k,j} | j=1,...,D\}$, a pixel-wise dense 3D cost volume can be built by differentiable warping on the learned image feature maps~\cite{yao2018mvsnet, wang2021patchmatchnet}.
% 3D cost volumes measure the 
% similarity between corresponding image
% features between reference image and
% source images.
To simplify the description, we ignore the stage index $k$ in the following formulation.
% as the operations in each search stage are similar in terms of $k$.

\par The input of MVS consists of relative camera rotation 
$\textbf{R}_{\textbf{F}_0 \rightarrow  \textbf{F}_i}$
and translation $\textbf{t}_{\textbf{F}_0 \rightarrow \textbf{F}_i}$
from a reference feature map $\mathbf{F}_0$ to 
a source feature map $\mathbf{F}_i$.
Their corresponding camera intrinsics $\textbf{K}_0, \textbf{K}_i$ are also known.
We first construct a set of 2-view cost volumes
$\{\textbf{V}_{i}\}_{i=1}^{N-1}$ from the $N-1$ source image feature maps
by differentiable warping and group-wise correlation \cite{guo2019group, xu2020pvsnet, wang2021patchmatchnet}.
Let $\textbf{p}$ be a pixel in $\mathbf{I}_0$,
$\textbf{p}'$ be the warped pixel 
of $\textbf{p}$ in the
source image $\mathbf{I}_i$ by the $j$-th depth hypothesis, \ie~$d_j$.
Then $\textbf{p}'$ can be computed by:
\begin{equation}\label{equ:warppixel}
\textbf{p}' = \textbf{K}_i \cdot (\textbf{R}_{\textbf{F}_0 \rightarrow  \textbf{F}_i} \cdot \textbf{K}_0^{-1} \cdot \textbf{p} \cdot d_j + \textbf{t}_{\textbf{F}_0 \rightarrow \textbf{F}_i}),
\end{equation}
% Let $\textbf{F}_0$ be the feature map of
% $I_0$, and $\textbf{F}_i$ be
% the feature map of source
% image $I_i$. 
where the feature maps $\mathbf{F}_0$ and $\mathbf{F}_i$ all have a channel dimension of $N_c$.
% They all have a channel dimension
% with $N_c$ channels.
Following \cite{guo2019group}, we divide the channels of the feature maps into $N_g$ groups along the channel dimension, and each feature group therefore has $N_c/N_g$ channels. Let $\textbf{F}_{i}^g$ be the $g$-th feature group of $\textbf{F}_i$.
Then we can compute the $i$-th cost volume $\textbf{V}_i$ from $\textbf{F}_i$ as follows:
\begin{equation} \label{equ:twoviewvolume}
\setlength\abovedisplayskip{3pt}
\setlength\belowdisplayskip{3pt}
    \textbf{V}_i(j, \textbf{p}, g) = \frac{N_g}{N_c} \langle \textbf{F}_{0}^g(\textbf{p}), \textbf{F}_{i}^g(\textbf{p}') \rangle.
\end{equation}
Where $\langle\cdot,\cdot \rangle$ denotes a correlation calculation by an inner product operation. The group-wise correlation allows us to more efficiently construct a full cost volume. After the construction of each 2-view cost volume, we apply several 3D CNN layers to predict a set of pixel-wise weight matrices $\{\textbf{W}_{i}\}_{i=1}^{N-1}$. Then we fuse these cost volumes into one cost volume $\textbf{V}$ via weighted fusion~\cite{xu2020pvsnet, wang2021patchmatchnet} with  $\{\textbf{W}_{i}\}_{i=1}^{N-1}$ as:

\begin{equation}
    \textbf{V}(j, \textbf{p}, g) = \frac{\sum_{i=1}^{N-1} \textbf{W}_{i}(\textbf{p}) \cdot \textbf{V}_i(j, \textbf{p}, g)}{\sum_{i=1}^{N-1} \textbf{W}_{i}(\textbf{p})}.
\end{equation}
The fused 3D cost volume $\textbf{V}$ is then regularized by a 3D UNet~\cite{ronneberger2015u, yao2018mvsnet}, which gradually reduces the channel size of $\textbf{V}$
to 1 and output a volume of size $D, H, W$, i.e. spatial size of the volume. Finally, a $\mathrm{Softmax(\cdot)}$ function is performed along the $D$ dimension
to produce a probability volume $\textbf{P}$ for computing the training loss and labels.

% We also know the camera parameters
% $\{\textbf{K}_i, \textbf{R}_i, \textbf{t}_i\}_{i=1}^{N-1}$
% of these images, where 
% are also known. 

\subsection{Binary Search for MVS} \label{sec:binarysearch}
% \par\noindent\textbf{Depth Hypotheses} 
Multi-view stereo networks~\cite{yao2018mvsnet, yao2019recurrent, gu2020cascade} typically densely sample depth hypotheses for each pixel to
construct 3D cost volumes, resulting in remarkably high memory footprint. To alleviate this issue, recent cascade-based MVS methods ~\cite{gu2020cascade, Cheng_2020_CVPR}
propose to construct cost volumes in a coarse-to-fine manner which reduces the memory usage to some extent. However, in each iterative stage, the sampling is still much dense, and thus the model efficiency is far less than optimal. 
% In our GBiNet, only $D=4$ depth hypotheses 
% are sampled for each each iteration, which largely
% decreases the memory usage of cost volumes.
\par In this work, we explore a reasonably optimal sampling strategy from a perspective of discrete search for highly-efficent MVS, and propose a binary search method (Bi-Net). Specifically,
instead of directly sampling depth values in the given depth range $R$,
we divide the current depth range into bins.
For the $k$-th search stage,
we divide the depth range into $2$ equal bins, \ie~$\{B_{k,j}| j=1,2\}$ with $B_{k,j}$ denoting a bin. The bin width of $B_{1,j}$ in the first stage is $R/2$. As we cannot directly use discrete bins for warping feature maps, we sample center points of the 2 bins to represent the depth hypotheses of bins, and then construct the cost volume and perform label prediction for the 2 bins.
Let the three edges from left to right of the 2 bins be $\{e_{k, m} | m=1, 2, 3\}$.
Then, the two edges of bin $B_{k,j}$ are $e_{k, j}$ and $e_{k, j+1}$. For instance, $e_{k,1}$ and $e_{k,2}$ are edges of $B_{k,1}$. Then
% These center points are used as depth hypotheses $\{d_{k,j} | k=1, j=1,2\}$
% for cost volume construction and label prediction. 
the depth hypothesis $d_{k,j}$ for the 2 bins can be computed as follows:

\begin{equation}
\label{equ:getcenterpoint}
% \vspace{-11pt}
d_{k, j} =  \frac{e_{k, j} + e_{k, j+1}}{2}, \ \  j=1,2.
% \vspace{-2pt}
\end{equation}
The predicted label of a depth hypothesis indicates whether
the true depth value is in the corresponding bin. In the $k$-th search stage, after the network outputs the probability volume $\textbf{P}$, we apply an $\mathrm{argmax}(\cdot)$ operation
along the $D$ dimension of $\textbf{P}$, which returns the label $j$ indicates that the true depth value is in the bin $B_{k,j}$. The new 2 bins at the $(k+1)$-th search stage can be further generated by dividing $B_{k,j}$ into two equal-width bins $B_{k+1,1}$ and $B_{k+1,2}$, and the corresponding three edges at this stage can be defined as:
\begin{equation}
    e_{k+1,1} = e_{k, j}; e_{k+1, 2} = \frac{e_{k, j} + e_{k, j+1}}{2}; e_{k+1, 3} =  e_{k, j+1}.
\end{equation}
Then new depth hypotheses are sampled from the center points of bins $B_{k+1,1}$ and $B_{k+1,2}$ for the $(k+1)$-th stage.
{The initial bin width in the proposed binary search is $R/2$ and in the $k$-th stage, the bin width is $R/ 2^{k}$.}

\par With our proposed binary search strategy, the depth dimension of the 3D cost volume can be decreased to 2, which pushes the cost volume size to an extremely low bound, and the memory footprint is dramatically decreased. In our experiments, the Binary Search for MVS achieves satisfactory results, outperforming several existing competitive methods, and the memory overhead of the whole MVS network becomes dominated by the 2D image encoder, no longer by the 3D cost volumes. However, as discussed in the Introduction,
the issue of the network classification error can cause unstable optimization and degraded accuracy.
% In order to address these issues and
% retain the memory advantage of Binary Search,
% we proposed Generalized Binary Search for MVS
% in next Section \ref{sec:generalizedbinarysearch}.

% For the $k$th iteration,
% the depth bins $\{B_{k,j} | j=1,...,D\}$ 
% are update from the previous iteration
% according to network prediction.

% In our multi-scale implementation,
% some iterations share the same 2D
% feature maps and spatial scales.
% When iteration $k$ and $k+1$ are
% not in the same scale, we upsample
% the prediction of iteration $k$
% by a factor of 2 and then update 
% the bins.
% We will introduce the updating of bins
% in Section \ref{sec:updatingandpadding}.

\begin{figure}[!t]
  \centering
   \includegraphics[width=\linewidth]{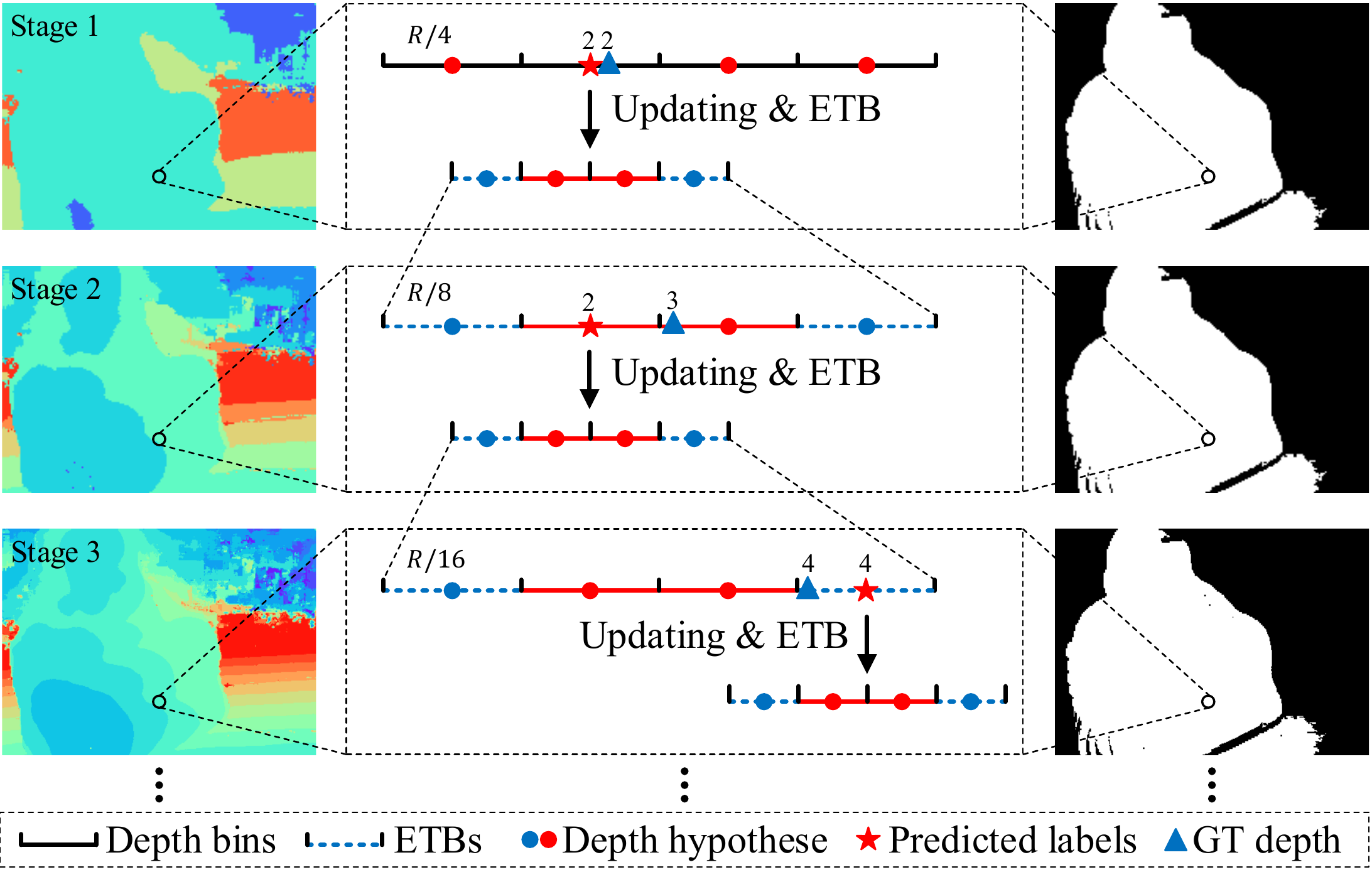}
   \vspace{-20pt}
   \caption{Illustration of Generalized Binary Search. We subdivide 
   the selected bin into two bins according to the label. 
   Then we pad Error Tolerance Bins (ETB) on both sides.
   We check the gradient mask of this pixel. Only if it is valid, 
   the loss could participate in back-propagation.}
   \label{fig:quadraticsearch}
   \vspace{-20pt}
\end{figure}

\subsection{Generalized Binary Search for MVS} \label{sec:generalizedbinarysearch}
To handle the error accumulation and the training issue in the proposed
Binary Search for MVS,
we extend it to a Generalized Binary Search for MVS. Specifically, we further design three effective mechanisms which are error tolerance bins, gradient-masked optimization and efficient gradient updating mechanism, making substantial improvement over the Binary Search method. 
% The main improvement is 
% achieved by Error Tolerance Bins and Gradient-Masked Optimization.
\par\noindent\textbf{Error Tolerance Bins.}
After obtaining the selected bin
$B_{k,j}$ in the $k$-th search stage, 
we first divide it into two new bins $B_{k+1,1}$ and $B_{k+1,2}$
for the next $(k+1)$-th stage, as shown in Fig.~\ref{fig:quadraticsearch}. To make the network have a certain capability of tolerating prediction errors, we propose to respectively add one small bins on the left side of $B_{k+1,1}$ and on the right side of $B_{k+1,2}$. This process is termed as Error Tolerance Bins (ETB).
More formally, given $D$ ($D$ is an even number and small enough) as the final number of bins, we pad $(D - 2)/2$ more bins to the two sides of the original two bins.
After the padding, $D$ new bins, \ie~ $\{B_{k+1,j} | j=1,...,D\}$, are obtained, as well as their corresponding bin edges, \ie~$\{e_{k+1, m} | m=1,...,D+1\}$.
We still sample the center points as 
the depth hypotheses $\{d_{k+1,j} | j=1,...,D\}$ from these bins with Eq.~\ref{equ:getcenterpoint}. The error tolerance bins
extend the sampling of depth hypotheses 
to a range out of the two original bins in the binary search, thus enabling the network to correct
the predictions and to reduce error
accumulation to some extent. Since the depth hypotheses number is now $D$, we also change the initialization of depth hypotheses in the first stage.
As the initial depth range $R$ in split
into $D$ bins, the initial bin width is $R/D$ and in the $k$-th stage, the bin width is $R/(D \times 2^{k-1})$.

\par In our network implementation, we pad only 1 ETB on both sides. This leads to a depth hypothesis number of 4, \ie~$D = 4$. In the experiments, we observe dramatically improved depth prediction accuracy while notably, the memory consumption can be the same level as the original binary search, as the memory is still dominated by the 2D image encoder. Fig.~\ref{fig:quadraticsearch}
shows a real example of our GBi-Net.
The hypothesis number $D$ is set to 4.
% For a pixel in the example image,
% at stage 1, the true depth is in $B_{1,2}$. The network also predicts a label of 2. Then the second bin is divided into 2 bins. We pad $(D - 2)/2=1$ bin before and after these 2 bins. In iteration 2, the 
% the true depth is in $B_{2,3}$.
% However, the network predicts
% a false label of 2. We still divide
% $B_{2,2}$ into 2 bins, 
% which no longer cover the true depth.
% This is a situation that binary search will accumulate errors.
With the error tolerance bins, the network can predict a correct label of 4 when the true depth is in $B_{3,4}$ at the $3$-th search stage, while the original binary search fails.

\par\noindent\textbf{Gradient-Masked Optimization.}
The proposed GBi-Net is trained in 
a supervised manner. The ground-truth labels are generated from the ground-truth depth map. In the $k$-th search stage, after we obtain the bins, we calculate which bin is occupied by the ground-truth
depth value. Then we can convert
the ground truth depth map into a ground truth occupancy volume $\textbf{G}$
with one-hot encoding, which is further used for loss calculation.
One problem in the iterative search is
that the ground-truth
depth values for some pixels may be out of the $D$ bins. 
In this situation, no valid labels
exist and the losses cannot be computed. 
This is a critical problem in network optimization. The coarse-to-fine methods typically leverage a continuous regression loss, while existing MVS methods with a discrete classification loss~\cite{yao2019recurrent} widely employ dense space discretization.

\par In our GBiNet, 
a designed mechanism to this problem
is computing a mask map for each stage, based on the bins and
ground-truth depth maps. 
If the ground-truth
depth of a pixel is in the current bins, the pixel is considered as valid.
% and only valid pixels are considered in the loss calculation.
Let the ground-truth depth for a pixel be $d_{gt}$ and the current bin edges be $\{e_{m} | m=1,...,D+1\}$, omitting the stage index for simplicity.
Then the pixel is valid only if:
\begin{equation}
  e_{1} \leq d_{gt} < e_{D+1}.
\end{equation}
Only the loss gradients from the valid pixels are used to update the parameters in the network. The gradients from all the invalid pixels are not accumulated. With this process, we can train both Bi-Net and GBi-net successfully, as clearly confirmed by our experimental results. The gradient-masked optimization is similar to the popular self-paced learning~\cite{sangineto2018self}, in which at the very beginning, the network only involves easy samples (\ie~easy pixels) in training, while with the optimization proceeds, the network can predict more accurate labels for hard pixels, and most pixels will eventually participate in the learning process.
% relatively hard pixels probably not participate in the
% network learning, \ie~only easy samples are utilized, however, with the training proceeds, the network can predict more accurate labels
% for hard pixels and they eventually participate in the learning process. 
As can be observed in Fig.~\ref{fig:Validpixelpercentagesub} in the
experiments, a large portion of pixels falls into the the current bins in our GBi-Net.

\subsection{Network Optimization}\label{sec:networkoptimization}

\noindent{\textbf{Loss Function.}}
Our loss function is a standard cross-entropy loss
that applies on the probability
volume $\textbf{P}$ and a ground truth occupancy volume
$\textbf{G}$.
A set of the valid pixels $\Omega_{\mathbf{q}}$ is first obtained by the valid mask map and then a mean loss
of all valid pixels is computed as follows:

\begin{equation}
    Loss = \sum_{\textbf{q} \in  \Omega_{\mathbf{q}}}\sum_{j=1}^{D}-\textbf{G}(j, \textbf{q})\log \textbf{P}(j, \textbf{q})
% \vspace{-10pt}
\end{equation}

% \noindent{\textbf{Memory-efficient Incremental Training.}}
\noindent{\textbf{Memory-efficient Training.}}
MVS methods with multiple stages~\cite{gu2020cascade, Cheng_2020_CVPR}
typically average the losses from all the stages and back-propagate the gradients together. Nevertheless, this training strategy consumes significant memory because of the gradients accumulation across different stages. In our GBi-Net, we train our network in a more memory-efficient way. Specifically, we compute the loss
and back-propagate the gradients immediately after each stage. The gradients are not accumulated across stages, and thus the maximum memory overhead does not exceed the stage with the largest scale. To make the training with multiple stages more stable, we first set the maximum number of search stages as 2, and gradually increase it as the epoch number increases.

% \par\noindent\textbf{Inference}

% \subsubsection{Depth Hypothesis Initialization}
% \label{sec:DepthHypothesisInitialization}

% \textbf{Initialization:} For the reference image, we know the
% In the first iteration,
% we divide the original depth range $R$
% Since our network perform quadratic search,
% we initialize 4 depth samples for the first iteration.
% Recently MVS networks usually directly sample 
% depth values uniformly or in an inverse-depth space.
% Different from this, we first divide the 
% initial depth range $R$ into for bins. 
% Our target is to predict which bin the truth
% depth is in. However, the bins is not single
% depth values. Warping operation could not
% be applied on the depth bins. Here we use
% one point to represent a bin. Practically,
% we use the mid-point of each bin to represent
% each bin. Then 4 depth samples is initialized 
% as the mod-points of the 4 bins.

% \subsubsection{Depth sample generation}
% bin and mid-points
% \subsubsection{Cost Volume Regularization}\label{sec:costvolumeregularization}

% \subsubsection{Updating and Padding}\label{sec:updatingandpadding}

% \subsubsection{Masked training}

% \subsubsection{Mask}

\begin{figure}[!t]
  \centering
   \includegraphics[width=\linewidth]{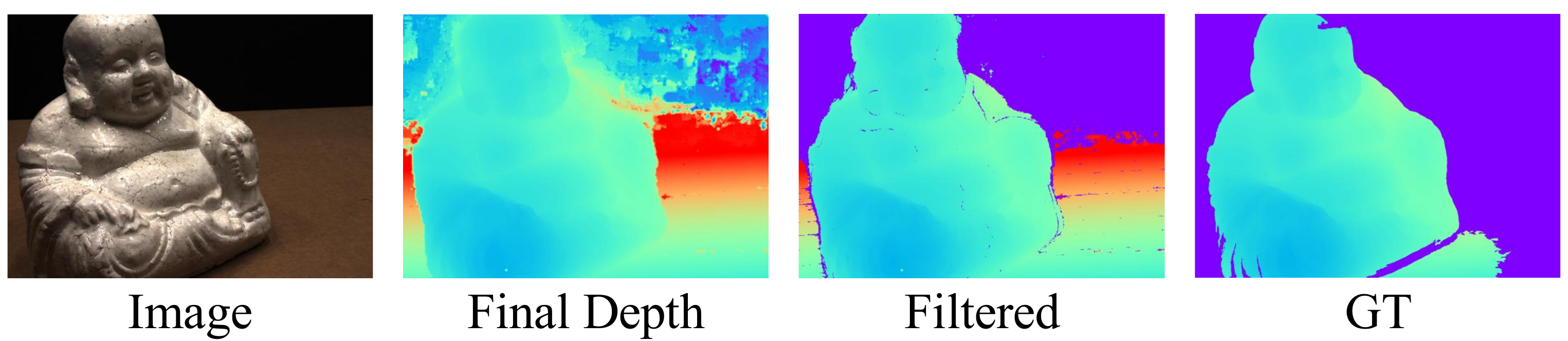}
   \vspace{-20pt}
   \caption{Visualization of the final and filtered depth maps.}
    \vspace{-10pt}
   \label{fig:depthmapexample}
\end{figure}

\begin{figure}[t]
  \centering
  \includegraphics[width=\linewidth]{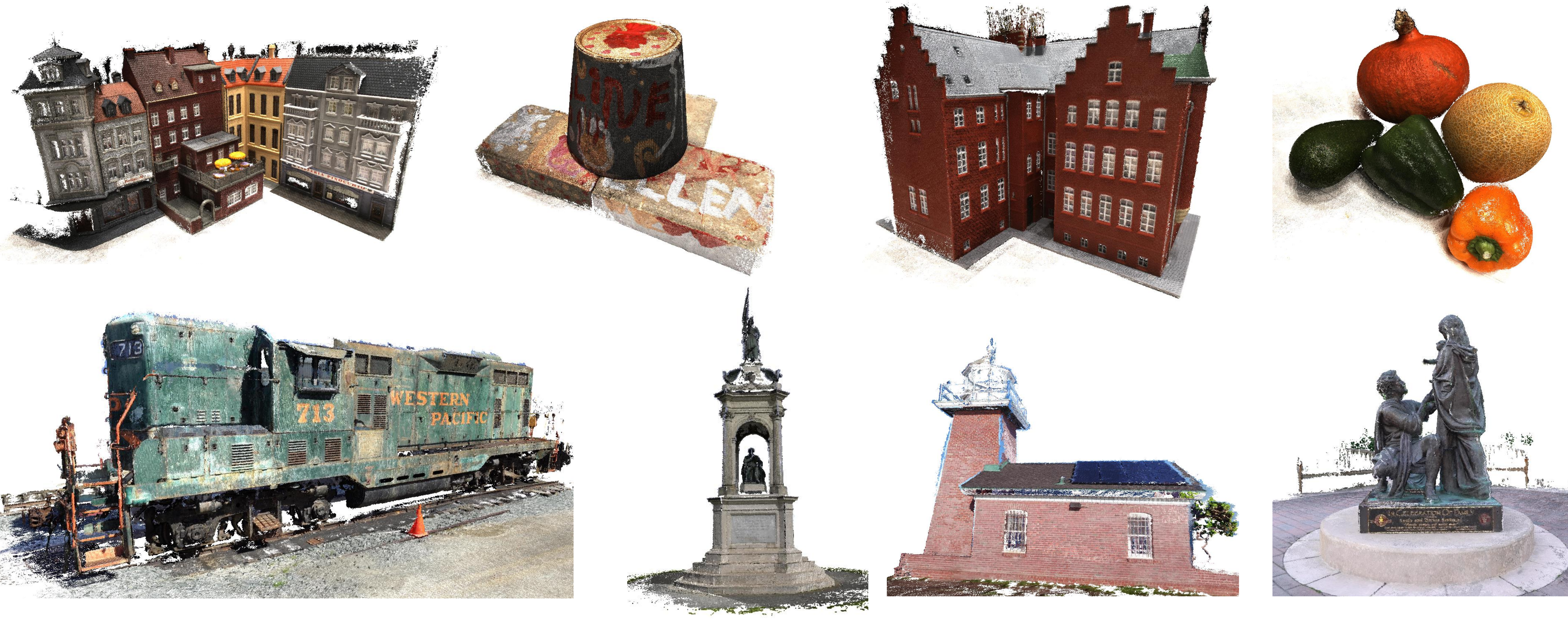}
  \vspace{-20pt}
  \caption{Point clouds examples of our method on DTU~\cite{aanaes2016large} and Tanks and Temples~\cite{knapitsch2017tanks}.}
  \label{fig:qualitative}
  \vspace{-10pt}
\end{figure}

% \subsection{Depth map Fusion}\label{sec:depthmapfusion}
% The final depth maps are generated from center points of the selected
% bins after the final stage.
% We then filter and fuse these depth maps into one point cloud.
% Similar to previous MVS methods \cite{yao2018mvsnet, yao2019recurrent},
% we consider photometric and the geometric
% consistencies for depth map filtering.
% For each pixel, we take the average of classification probabilities 
% from different stages and use the average value as photometric consistency.
% In practice, we take the average probability of first 6 stages.
% The geometric consistency is similar to \cite{yao2018mvsnet} measuring
% the depth consistency among multiple views.
% After filtering, we fuse them into a point cloud. Fig.~\ref{fig:depthmapexample} shows a visualization result of our
% method.

% \subsection{Implementation Details}
% Detailed Network Architecture
% loss function

% training

\section{Experiments}
\label{sec:experiments}
% We benchmark our GBiNet on multiple datasets
% and perform several ablation studies.
% , such as DTU \cite{aanaes2016large}, Tanks and Temples\cite{knapitsch2017tanks}.
\subsection{Datasets}
% \noindent{\textbf{DTU Dataset.}}  
The \textbf{DTU} dataset~\cite{aanaes2016large} 
is an indoor dataset with multi-view images and camera poses. We Follow MVSNet~\cite{yao2018mvsnet} for dividing training and testing set. There are 27097 training samples in total. 
The \textbf{BlendedMVS} dataset~\cite{yao2020blendedmvs} 
is a large-scale dataset with indoor and outdoor scenes. Following~\cite{zhang2020visibilityaware,Wei_2021_ICCV,ma2021epp}, we only use this dataset for training. There are 16904 training samples in total. 
% We perform online data augmentation via modifying the brightness and contrast of input images randomly, similar to~\cite{yao2020blendedmvs}. 
\textbf{Tanks and Temples}~\cite{knapitsch2017tanks} is a large-scale dataset with various outdoor scenes. It contains Intermediate subset and Advanced subset. The evaluation on this benchmark is conducted online by submitting generated point clouds to the official website. 
% The evaluation results could be found in the leader board
% of the website.

\begin{table}[t]
  \centering
  \vspace{-6pt}
  \caption{Point cloud evaluation results on DTU~\cite{aanaes2016large}. The lower is better for Accuracy (Acc.), Completeness (Comp.), and Overall. The best result is highlighted in bold and the second in italic bold. * denotes ours without using random cropping data augmentation.}
  \vspace{-8pt}
    \resizebox{0.95\linewidth}{!}{\begin{tabular}{lcccc}
    \toprule
    Method & Acc.$\downarrow$ & Comp.$\downarrow$ & Overall$\downarrow$ & Mem. (MB)\\
    \midrule
    % Camp \cite{campbell2008using}  & 0.835 & 0.554 & 0.695 \\
    % Furu \cite{furukawa2009accurate} & 0.613 & 0.941 & 0.777 \\
    Tola \cite{tola2012efficient} & 0.342 & 1.190  & 0.766 & - \\
    Gipuma \cite{galliani2015massively} & \textbf{0.283} & 0.873 & 0.578 & - \\
    % SurfaceNet \cite{ji2017surfacenet} & 0.450  & 1.04  & 0.745 \\
    MVSNet \cite{yao2018mvsnet} & 0.396 & 0.527 & 0.462 & 9384\\
    R-MVSNet \cite{yao2019recurrent} & 0.383 & 0.452 & 0.417 & - \\
    CIDER \cite{xu2020learning} & 0.417 & 0.437 & 0.427 & - \\
    % P-MVSNet \cite{luo2019p} & 0.406 & 0.434 & 0.420 \\
    Point-MVSNet \cite{chen2019point} & 0.342 & 0.411 & 0.376 & - \\
    % Fast-MVSNet \cite{Yu_2020_CVPR} & 0.336 & 0.403 & 0.370 \\
    CasMVSNet \cite{gu2020cascade} & 0.325 & 0.385 & 0.355 & 4591 \\
    UCS-Net \cite{Cheng_2020_CVPR} & 0.338 & 0.349 & {0.344} & 4057 \\
    CVP-MVSNet \cite{yang2020cost} & \textbf{\textit{0.296}} & 0.406 & 0.351 & - \\
    Vis-MVSNet \cite{zhang2020visibilityaware} & 0.369 & 0.361 & 0.365 & 4775 \\
    PatchmatchNet \cite{wang2021patchmatchnet} & 0.427 & {0.277} & 0.352 & \textbf{1629} \\
    AA-RMVSNet \cite{Wei_2021_ICCV} & 0.376 & 0.339 & 0.357 & 11973 \\
    EPP-MVSNet \cite{ma2021epp} & 0.413 & 0.296 & 0.355 & -\\
    % PVSNet & 0.337 & 0.315 & 0.326 \\
    \midrule
    \midrule
    \textbf{Bi-Net (ours)} & 0.360  & 0.360  & 0.360 & \textbf{\textit{2108}}  \\
    \textbf{GBi-Net* (ours)} & 0.327 & \textbf{\textit{0.268}} & \textbf{\textit{0.298}} & \textbf{\textit{2108}} \\
    \textbf{GBi-Net (ours)} & 0.315 & \textbf{0.262} & \textbf{0.289} & \textbf{\textit{2108}} \\
    \bottomrule
    \end{tabular}}%
  \label{tab:dtuevaluateresult}%
  \vspace{-5pt}
\end{table}%

\begin{table}[t]
  \centering
  \caption{Depth map evaluation results in terms of accuracy, and the memory consumption on DTU~\cite{aanaes2016large}. The unit of these
  thresholds are all in millimeters. 
%   The score is the percentage of pixels whose absolute depth error are less than the threshold. 
  The higher is better.}
    \vspace{-8pt}
    \resizebox{\linewidth}{!}{\begin{tabular}{cccccc}
  \toprule
    Method & $<$0.125$\uparrow$ & $<$0.25$\uparrow$ & $<$0.5$\uparrow$ & $<$1$\uparrow$ & Mem. (MB)\\
    \midrule
    MVSNet \cite{yao2018mvsnet} & 8.539 & 16.85 & 32.02 & 53.61 & 9384 \\
    CasMVSNet \cite{gu2020cascade} & 10.13 & 19.88 & 37.04 & 59.4  & 4591\\
    Patchmatchnet \cite{wang2021patchmatchnet} & 8.113 & 16.05 & 30.77 & 52.69 & 1629\\
    % GBi-Net-resize & \underline{11.75} & \underline{22.92} & \underline{41.62} & \underline{60.99} & 2108\\
    \midrule
     \midrule
    \textbf{GBi-Net (ours)} & \textbf{12.77} & \textbf{24.89} & \textbf{45.1}  & \textbf{65.94} & 2108\\
    \bottomrule
    \end{tabular}}%
    \label{tab:thresratiocompare}%
    \vspace{-15pt}
\end{table}%

\begin{table*}[t]
  \centering
  \caption{Point cloud evaluation results on the Advanced and Intermediate subsets of Tanks and Temples dataset \cite{knapitsch2017tanks}.
  Higher scores are better. The Mean is the average score of all scenes.}
  \vspace{-8pt}
    \resizebox{\linewidth}{!}{\begin{tabular}{lccccccc|ccccccccc}
    \toprule
    & \multicolumn{7}{c}{Advanced} & \multicolumn{9}{|c}{Intermediate} \\
    \midrule
    Method & Mean  & Aud.  & Bal.  & Cou.  & Mus.  & Pal.  & Tem.  & Mean  & Fam.  & Fra.  & Hor.  & Lig.  & M60   & Pan.  & Pla.  & Tra. \\
    \midrule
    MVSNet~\cite{yao2018mvsnet} &   -    &   -    &    -   &   -    &    -   &   -    &     -  & 43.48 & 55.99 & 28.55 & 25.07 & 50.79 & 53.96 & 50.86 & 47.90  & 34.69 \\
    Point-MVSNet~\cite{chen2019point} &   -    &    -   &    -   &    -   &   -    &   -    &  -     & 48.27 & 61.79 & 41.15 & 34.20  & 50.79 & 51.97 & 50.85 & 52.38 & 43.06 \\
    UCSNet~\cite{Cheng_2020_CVPR}&   -    &    -   &    -   &  -     &   -    &    -   &   -    & 54.83 & 76.09 & 53.16 & 43.03 & 54.00    & 55.60  & 51.49 & 57.38 & 47.89 \\
    CasMVSNet~\cite{gu2020cascade}& 31.12 & 19.81 & 38.46 & 29.10  & 43.87 & 27.36 & 28.11 & 56.42 & 76.36 & 58.45 & 46.20  & 55.53 & 56.11 & 54.02 & 58.17 & 46.56 \\
    PatchmatchNet~\cite{wang2021patchmatchnet} & 32.31 & 23.69 & 37.73 & 30.04 & 41.80  & 28.31 & 32.29 & 53.15 & 66.99 & 52.64 & 43.24 & 54.87 & 52.87 & 49.54 & 54.21 & 50.81 \\
    BP-MVSNet~\cite{sormann2020bp} & 31.35 & 20.44 & 35.87 & 29.63 & 43.33 & 27.93 & 30.91 & 57.60  & 77.31 & 60.90  & 47.89 & 58.26 & 56.00    & 51.54 & 58.47 & 50.41 \\
    Vis-MVSNet~\cite{zhang2020visibilityaware} & 33.78 & 20.79 & 38.77 & 32.45 & 44.20  & 28.73 & 37.70  & 60.03 & 77.40  & 60.23 & 47.07 & 63.44 & 62.21 & 57.28 & 60.54 & 52.07 \\
    AA-RMVSNet~\cite{Wei_2021_ICCV} & 33.53 & 20.96 & 40.15 & 32.05 & 46.01 & 29.28 & 32.71 & 61.51 & 77.77 & 59.53 & 51.53 & \textbf{64.02} & \textbf{64.05} & 59.47 & 60.85 & 54.90 \\
    EPP-MVSNet~\cite{ma2021epp} & 35.72 & 21.28 & 39.74 & 35.34 & \textbf{49.21} & 30.00    & \textbf{38.75} & \textbf{61.68} & 77.86 & 60.54 & \textbf{52.96} & 62.33 & 61.69 & \textbf{60.34} & \textbf{62.44} & \textbf{55.30}\\
    \midrule
    \midrule
    \textbf{Bi-Net (ours)} & 32.03 & 21.97 & 37.59 & 31.63 & 44.81 & 27.92 & 28.30  & 53.41 & 74.93 & 54.37 & 45.09 & 51.86 & 49.09 & 49.56 & 55.76 & 46.67 \\
    \textbf{GBi-Net (ours)} & \textbf{37.32} & \textbf{29.77} & \textbf{42.12} & \textbf{36.30}  & 47.69 & \textbf{31.11} & 36.93 & 61.42 & \textbf{79.77} & \textbf{67.69} & 51.81 & 61.25 & 60.37 & 55.87 & 60.67 & 53.89 \\
    \bottomrule
    \end{tabular}}%
  \label{tab:tanksanvancedintermediate}%
  \vspace{-15pt}
\end{table*}%

\subsection{Implementation Details} \label{sec:ImplementationDetails}
\noindent{\textbf{Training Details.}} 
The proposed GBi-Net is trained on the DTU dataset for DTU benchmarking and
trained on BlendedMVS dataset
for Tanks and Temples benchmarking, 
following~\cite{zhang2020visibilityaware,Wei_2021_ICCV,ma2021epp}. We use the high-resolution
DTU data provided by the open source 
code of MVSNet \cite{yao2018mvsnet}.
The original image size is $1200 \times 1600$.
% In MVSNet ~\cite{yao2018mvsnet}
% Existing MVS methods~\cite{yao2018mvsnet, gu2020cascade, wang2021patchmatchnet} usually follow the 
We first crop the input images into $1024 \times 1280$
following MVSNet~\cite{yao2018mvsnet}.
Different from MVSNet~\cite{yao2018mvsnet} that directly
downscale the image to $512 \times 640$,
we propose an online random cropping data augmentation.
We randomly crop 
images of $512 \times 640$ from images of $1024 \times 1280$.
The motivation is 
that cropping smaller images from larger images
could help to learn better features for
larger image scales without increasing the
training overhead.
% The number of input images is set to $N = 5$ 
When training on BlendedMVS dataset,
we use the original 
resolution of $576 \times 768$.
For all the training, $N = 5$ input 
images are used, \ie~1 reference 
image and 4 source images. We adopt the robust training strategy proposed in 
Patchmatchnet~\cite{wang2021patchmatchnet} for better learning of pixel-wise weights.
The maximum stage number is set to 8.
For every 2 stages, we share the same feature map scale and the 3D-CNN network parameters. The whole network is optimized by Adam optimizer in Pytorch
for 16 epochs with an initial learning rate of 0.0001, which is down-scaled by a factor of 2 after 10, 12, and 14 epochs. 
The total training batch size is 4 on two NVIDIA RTX 3090 GPUs.

\noindent{\textbf{Testing Details.}}
The model trained on DTU training set is used for testing on DTU testing set.
The input image number $N$ is set to 5, 
each with a resolution of $1152 \times 1600$. It takes 0.61 seconds for each testing sample. The model trained on BlendedMVS dataset is used for testing on
Tanks and Temples dataset. The image sizes are set to $1024 \times 1920$ or $1024 \times 2048$ to make the images divisible by 64. The input image number $N$ is set to 7.
All the testings are conducted on an NVIDIA RTX 3090 GPU. 
We then filter and fuse depth maps of a scene into one point cloud,
% by photometric and the geometric consistencies. The geometric
% consistency is defined similar to that of MVSNet \cite{yao2018mvsnet}.
% The photometric is defined from the average of classification probabilities,
details in the supplemental file.
Fig.~\ref{fig:depthmapexample} and Fig.~\ref{fig:qualitative} are visualizations of
depth maps and point clouds of our method.

\subsection{Benchmark Performance}
\par\noindent{\textbf{Overall Evaluation on DTU Dataset.}} 
We evaluate the results on the DTU testing set by two types of metrics. 
The first type of metric evaluates point clouds using official evaluation scripts of DTU~\cite{aanaes2016large}. It compares the distance
between ground-truth point clouds and the produced point clouds.
% The input images are set to resolution 1600 × 1200 
% with number of views N = 5. 
% The initial depth range for sampling hypotheses is fixed to [425mm,935mm], 
% according to DTU dataset. 
The state-of-the-art comparison results are shown in Table~\ref{tab:dtuevaluateresult}. 
Our two models, GBi-Net and GBi-Net* both significantly improved
the best performance on the completeness and the most important overall score (lower is better for both metrics).
% Note that Gipuma \cite{galliani2015massively} 
% is a traditional method.
% Compared with UCSNet~\cite{Cheng_2020_CVPR}, 
Our best model improves the overall
score from \textbf{0.344} of UCSNet~\cite{Cheng_2020_CVPR} to \textbf{0.289}, while the memory is reduced by $48\%$. Note that our Bi-Net, \ie~the proposed Binary Search
Network can also achieve comparable results to other dense search methods, clearly showing its effectiveness. 
% On the memory overhead, our Bi-Net and GBi-Nets are comparable to the most efficient PatchmatchNet~\cite{wang2021patchmatchnet} while our depth performance is remarkably better than PatchmatchNet (0.289 vs.~0.352 on the overall metric). 
The second type of metric directly evaluates the accuracy of the predicted depth maps. The depth ranges of DTU dataset are all 510 millimeters. 
% $935-425 = 510mm$. 
Thus, we compute the depth accuracy, which counts the percentage of pixels whose absolute depth errors are less
than a threshold, and 4 thresholds are considered in the evaluation (\ie~0.125, 0.25, 0.5, 1, with millimeters as a unit).
Compared to the depth range of 510 mm, these thresholds are extremely tight and challenging. The results of this type of metric are shown in Table~\ref{tab:thresratiocompare}.
Our GBi-Net also obtains the best results on all the thresholds.
The quality of depth maps also explains
our best performance on point cloud evaluation.

% \begin{table}[htbp]
%   \centering
%   \caption{Depth map evaluation results in terms of accuracy, and the memory consumption on DTU. The unit of these
%   thresholds are all in millimeters. 
% %   The score is the percentage of pixels whose absolute depth error are less than the threshold. 
%   The higher is better.}
%     \vspace{-8pt}
%     \resizebox{\linewidth}{!}{\begin{tabular}{cccccc}
%   \toprule
%     Method & $<$0.125$\uparrow$ & $<$0.25$\uparrow$ & $<$0.5$\uparrow$ & $<$1$\uparrow$ & Mem. (MB)\\
%     \midrule
%     MVSNet \cite{yao2018mvsnet} & 8.539 & 16.85 & 32.02 & 53.61 & 9384 \\
%     CasMVSNet \cite{gu2020cascade} & 10.13 & 19.88 & 37.04 & 59.4  & 4591\\
%     Patchmatch \cite{wang2021patchmatchnet} & 8.113 & 16.05 & 30.77 & 52.69 & 1629\\
%     % GBi-Net-resize & \underline{11.75} & \underline{22.92} & \underline{41.62} & \underline{60.99} & 2108\\
%     \midrule
%      \midrule
%     \textbf{GBi-Net (ours)} & \textbf{12.77} & \textbf{24.89} & \textbf{45.1}  & \textbf{65.94} & 2108\\
%     \bottomrule
%     \end{tabular}}%
%     \label{tab:thresratiocompare}%
% \end{table}%

\par\noindent{\textbf{Overall Evaluation on Tanks and Temples.}} 
We train the proposed Bi-Net and GBi-Net on BlendedMVS~\cite{yao2020blendedmvs}, 
and testing on Tanks and Temples dataset. 
We compare our method to state-of-the-art methods. Table~\ref{tab:tanksanvancedintermediate} shows results on both the Advanced subset and the Intermediate subset. Our GBi-Net achieves the best mean score of
37.32 (higher is better) on Advanced subset compared to all the competitors, and it performs the best on 4 out of the overall 6 scenes.
Note that the Advanced subset contains different large-scale outdoor scenes. The results can fully confirm the effectiveness of our method.
% showing the robustness of our method.
Table~\ref{tab:tanksanvancedintermediate} also shows the evaluation results on the
Intermediate subset. Our GBi-Net obtains highly comparable results to the state-of-the-art. Notably, with significantly less memory, our mean score is only 0.09 lower than AA-RMVSNet~\cite{Wei_2021_ICCV} and 0.26 lower than EPP-MVSNet~\cite{ma2021epp}. Moreover, we also obtain state-of-the-art scores on the \textit{Family} and \textit{Francis} scenes. Our binary search model Bi-Net also achieves satisfactory performance on both subsets.
% In order to test the generalization of our model, we test our method on Tanks and Temples dataset. The result of our proposed method on intermediate subset is demonstrated in Tab.\ref{tab:tanksintermediate}.
% The evaluation results are divided into two independent sections according to the training set of data. When our model is tested using DTU traning set, we achieve competitive performance compared to state-of-the-art method,such as CasMVSNet \cite{gu2020cascade}. And when we use BlendedMVS dataset for training, the performance of our model is not inferior to the previous ones, especially the reconstruction result on Family and Francis scenes outperforms existing methods with a significant margin. And the performance on advanced subset is shown in Tab.\ref{tab:tanksadvanced}. Since all the model utilized BlendedMVS dataset for training, we just simply demonstrate them together. In contrast to the existing methods, our model ranks $1^{st}$ on Tanks and Temples leaderboard with mean F1-score 37.32. Our method significantly improves robustness and generalization for various scenarios, such as Auditorium, Ballroom, Courtroom and Palace. 
% % Fig.\ref{} 
% visualizes qualitative results, which shows eminent performance in challenging areas, such as low textured planes, occluded boundaries and tiny objects which benefited from quadratic search optimizing strategy.
Our \textbf{anonymous} evaluation results on the leaderboard~\cite{tanksandtemplesleader}
are named as {Bi-Net} and {GBi-Net}.

\par\noindent{\textbf{Memory Efficiency Comparison.}} 
We compare the memory overhead with several previous best-performing learning-based MVS methods~\cite{yao2018mvsnet,Cheng_2020_CVPR,gu2020cascade,wang2021patchmatchnet,Yu_2020_CVPR,zhang2020visibilityaware,Wei_2021_ICCV} on the DTU testing set. The memory usage evaluation is conducted with an image 
size of $1152 \times 1600$. We use pytorch functions\footnote{max\_memory\_allocated
and reset\_peak\_memory\_stats} to measure the peak allocated memory usage of all the methods.
Fig.~\ref{fig:motivation}a shows a comparison of the methods regarding memory usage and  reconstruction error.
Our GBi-Net shows a great improvement in the reconstruction quality while using much less memory. More specifically, the memory footprint is reduced by 77.5\% compared to MVSNet~\cite{yao2018mvsnet}, by 54.1\% compared to CasMVSNet~\cite{gu2020cascade}, by 82.4\% to AA-RMVSNet~\cite{Wei_2021_ICCV}, and by 55.9\% to VisMVSNet~\cite{zhang2020visibilityaware}. Although the memory of our method is slightly $479$MB larger than Patchmatchnet~\cite{wang2021patchmatchnet}, 
shown in Table~\ref{tab:dtuevaluateresult} and~\ref{tab:thresratiocompare},
our method significantly outperforms it in both the point cloud (0.289 vs.~0.352) 
and depth map (12.77 vs.~8.113) evaluation by a large margin.

% \begin{table}[htbp]
%   \centering
%   \caption{Point cloud evaluation results on the Advanced subset of Tanks and Temples dataset.
%   Higher scores are better. The Mean is the average score of all scenes.}
%     \resizebox{\linewidth}{!}{\begin{tabular}{cccccccc}
%     \toprule
%     Method & Mean  & Audi. & Ball. & Court. & Mus. & Pal. & Tem. \\
%     \midrule
%     % BP-MVSNet \cite{sormann2020bp} & 31.35 & 20.44 & 35.87 & 29.63 & 43.33 & 27.93 & 30.91 \\
%     CasMVSNet \cite{gu2020cascade} & 31.12 & 19.81 & 38.46 & 29.10  & 43.87 & 27.36 & 28.11 \\
%     PatchmatchNet \cite{wang2021patchmatchnet} & 32.31 & 23.69 & 37.73 & 30.04 & 41.80  & 28.31 & 32.29 \\
%     EPP-MVSNet \cite{ma2021epp} & 35.72 & 21.28 & 39.74 & 35.34 & \textbf{49.21} & 30.00    & \textbf{38.75} \\
%     % Ours  & 34.63 & 26.05 & 39.36 & 34.88 & 47.17 & 28.3  & 32.03 \\
%     GBi-Net w/o ETB & 32.03 & 21.97 & 37.59 & 31.63 & 44.81 & 27.92 & 28.3 \\
%     GBi-Net & \textbf{37.32} & \textbf{29.77} & \textbf{42.12} & \textbf{36.30}  & 47.69 & \textbf{31.11} & 36.93 \\

%     \bottomrule
%     \end{tabular}}%
%   \label{tab:tanksadvanced}%
% \end{table}%

\subsection{Model Analysis}
% In this section, we conduct several ablation study to analyse
% the efficiency and accuracy of our method.
\noindent{\textbf{Effect of Different Search Strategies.}}
We first conduct a direct comparison on different search strategies as shown in Table~\ref{tab:searchablation}, including dense linear search by regression (\ie~Dense LS), dense coarse-to-fine search by regression (\ie~Dense C2F), and our Bi-Net and GBi-Net search by discrete classification. In this comparison, Bi-Net and GBi-Net are trained without using the random cropping data augmentation described in Sec.~\ref{sec:ImplementationDetails}. As we can observe from Table~\ref{tab:searchablation}, our binary search networks achieve significantly better results than Dense LS and Dense C2F on both the depth performance and the memory footprint, fully confirming the effectiveness of the proposed methods.

\begin{table}[t]
  \centering
  \caption{Performance comparison of different search strategies for MVS on DTU \cite{aanaes2016large}. LS indicates linear search,
%   \ie~MVSNet~\cite{yao2018mvsnet}. 
  and C2F indicates coarse-to-fine search. Both of them perform search via regression in a dense manner.
%   \ie~CasMVSNet~\cite{gu2020cascade}. 
%  Cla. denotes classification. 
%   GBi-Net (Cla.) are our Generalized Binary Search Net with classification. 
%   Bi-Net and GBi-Net are trained without the random cropping data augmentation described in Sec.~\ref{sec:ImplementationDetails}.
  }
  \vspace{-6pt}
    \resizebox{\linewidth}{!}{\begin{tabular}{lcccc}
    \toprule
    Method & Acc. $\downarrow$ & Comp. $\downarrow$ & Overall $\downarrow$ & Mem (MB) $\downarrow$\\
    \midrule
    Dense LS (Regression) & 0.396 & 0.527 & 0.462 & 9384 \\
    Dense C2F (Regression) & 0.325 & 0.385 & 0.355 & 4591 \\
    Bi-Net (Classification) & 0.360  & 0.360  & 0.360  & \textbf{2108} \\
    GBi-Net (Classification) & \textbf{0.327} & \textbf{0.268} & \textbf{0.298} & \textbf{2108} \\
    % GBi-Net (4 ETBs, Cla.) & \textbf{0.318} & \textbf{0.282} & \textbf{0.300}   & 2438 \\
    \bottomrule
    \end{tabular}}%
  \label{tab:searchablation}%
  \vspace{-10pt}
\end{table}%

\begin{figure}[htbp]
	\centering
	\begin{subfigure}[b]{0.49\linewidth}
		\centering
		\includegraphics[width=\linewidth]{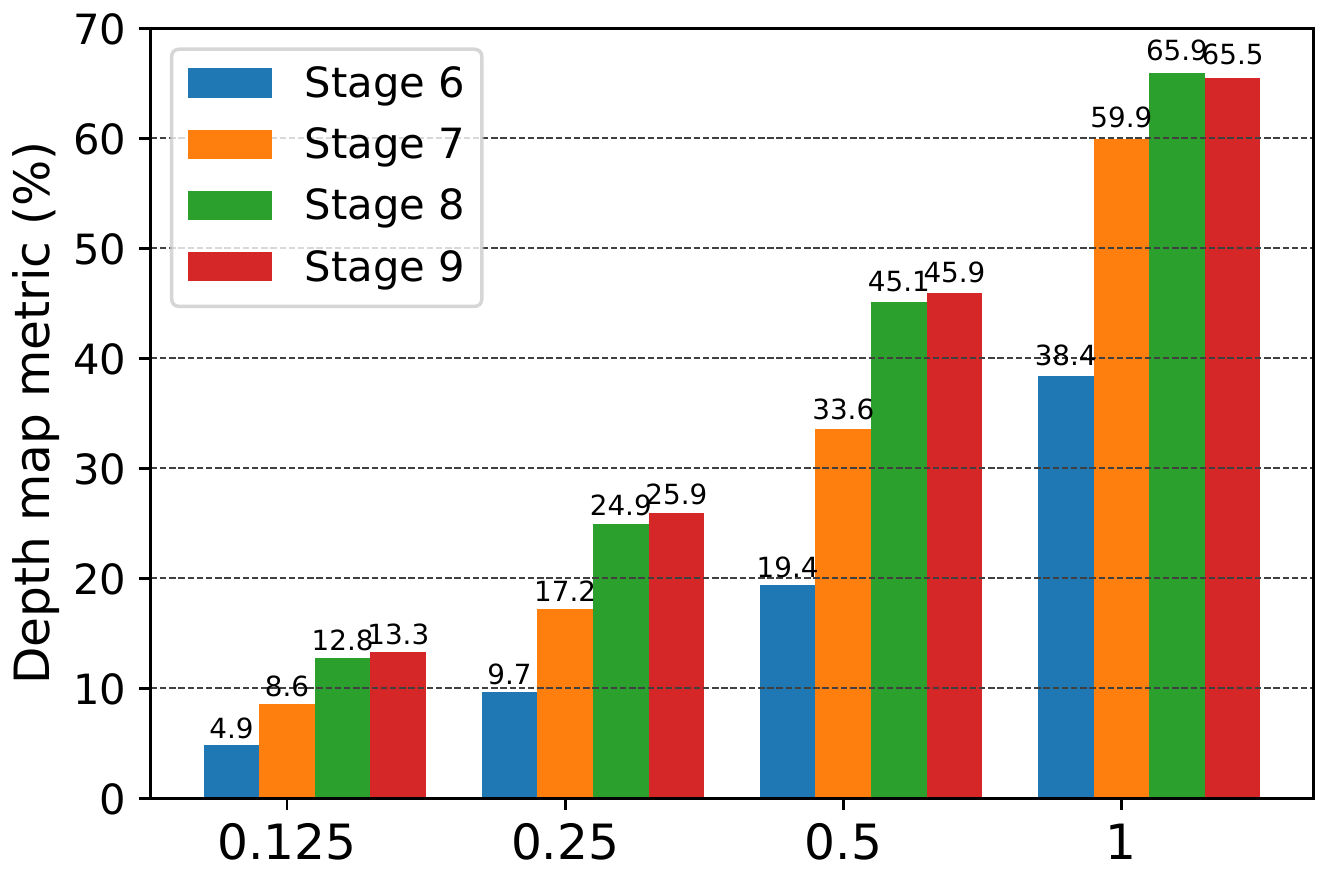}
		\caption[]%
		{{Depth map evaluation}}    
		\label{fig:Depthmapevaluationsub}
	\end{subfigure}
% 	\hfill
	\begin{subfigure}[b]{0.49\linewidth}  
		\centering 
		\includegraphics[width=\linewidth]{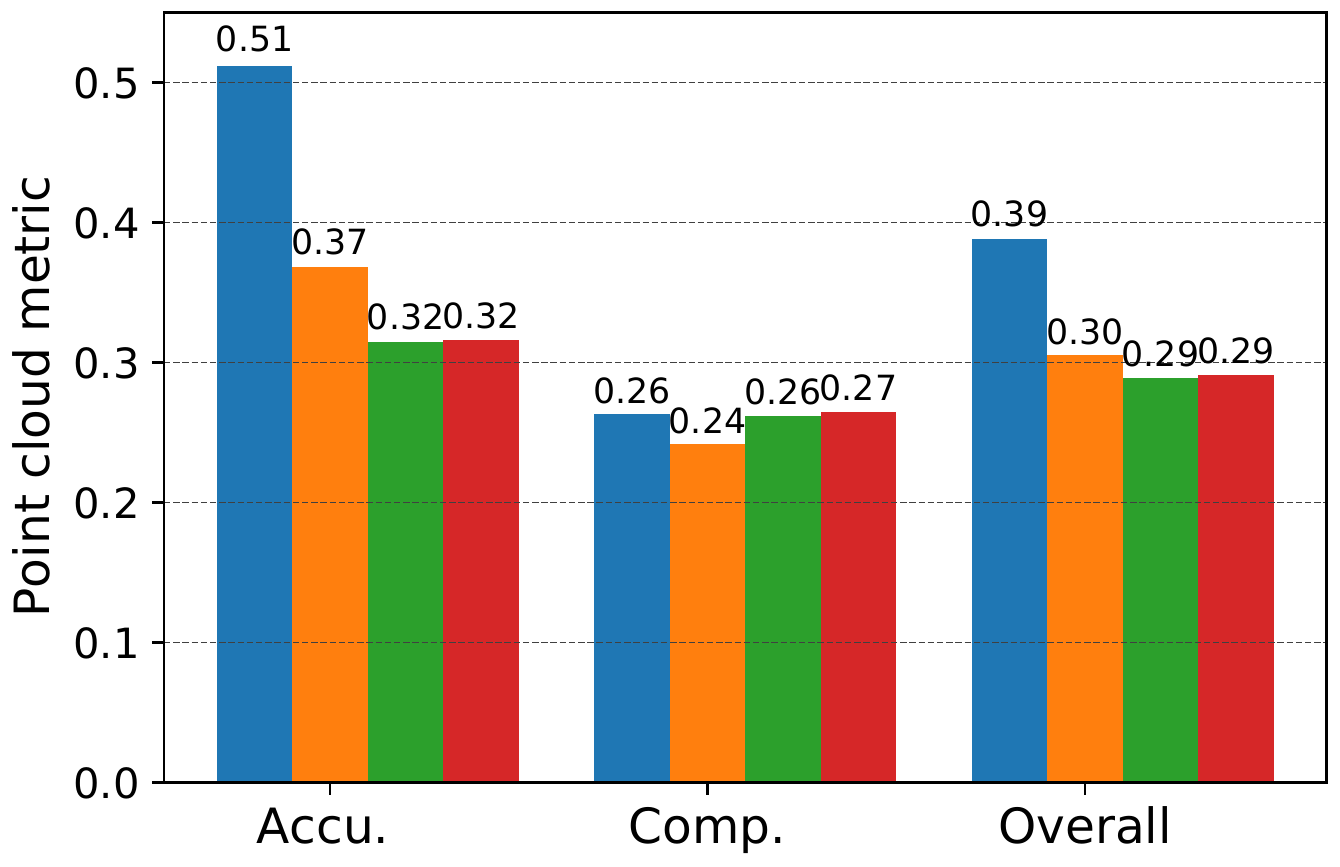}
		\caption[]%
		{{Point cloud evaluation}}    
		\label{fig:Pointcloudevaluationsub}
	\end{subfigure}
% 	\vskip\baselineskip
	\begin{subfigure}[b]{0.49\linewidth}   
		\centering 
		\includegraphics[width=\linewidth]{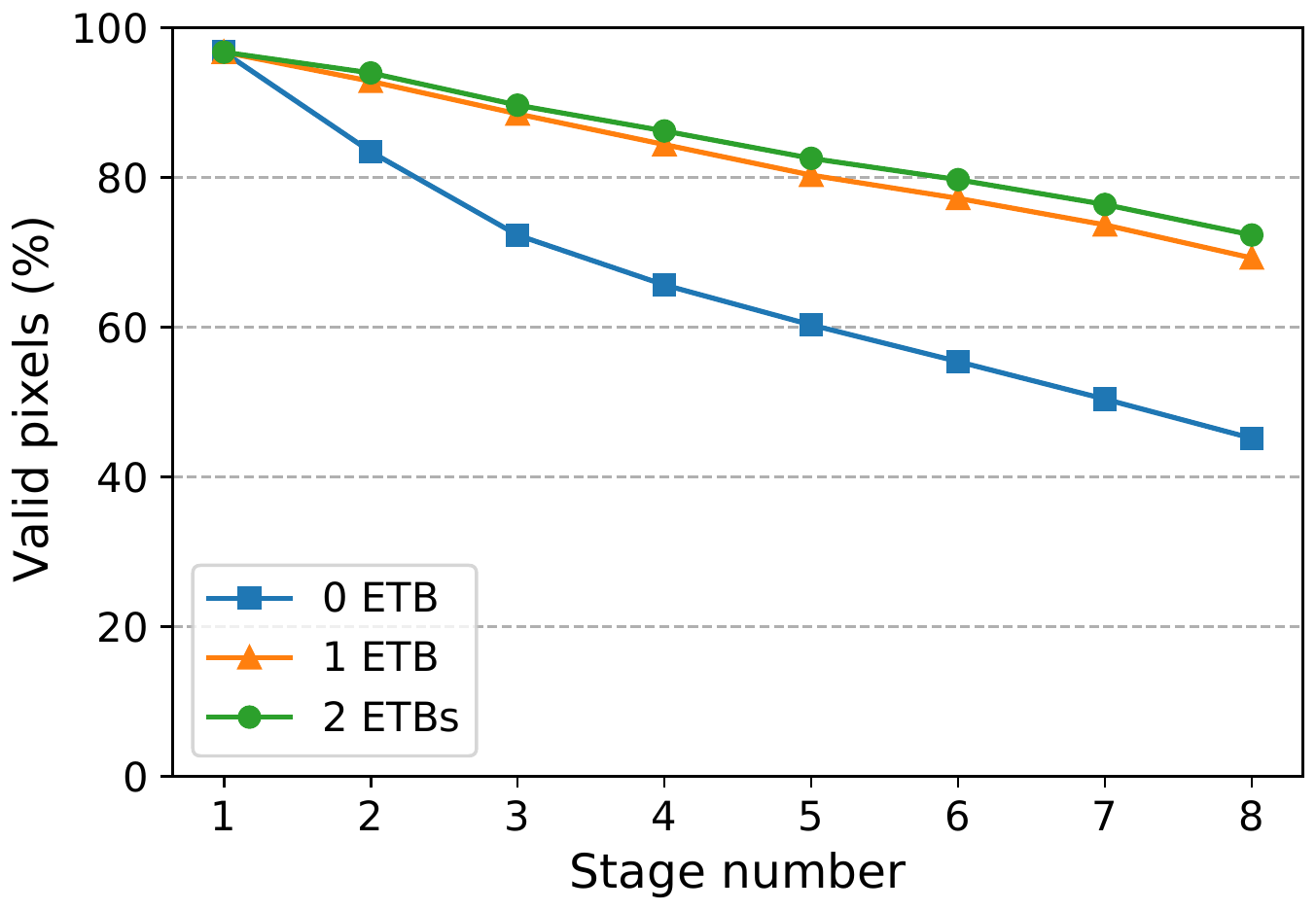}
		\caption[]%
		{{Valid pixel percentage}}    
		\label{fig:Validpixelpercentagesub}
	\end{subfigure}
% 	\hfill
	\begin{subfigure}[b]{0.49\linewidth}   
		\centering 
		\includegraphics[width=\linewidth]{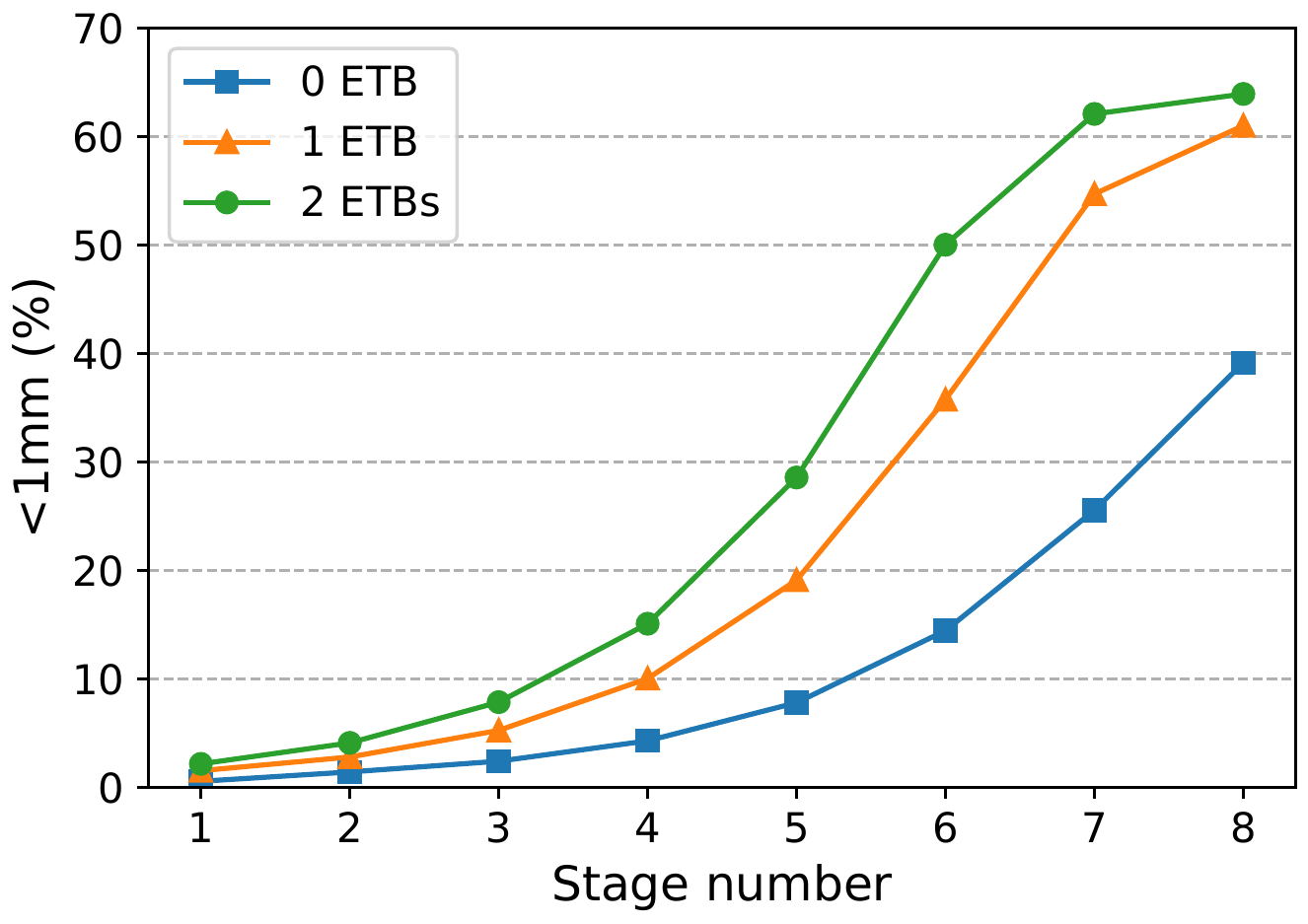}
		\caption[]%
		{{$<$1mm percentage}}    
		\label{fig:1mmpercentagesub}
	\end{subfigure}
	\vspace{-10pt}
	\caption[]
	{\small (a) and (b) are evaluation results of different stage numbers. (c) and (d) are valid pixel percentage and $<$1mm percentage of different stage numbers of models with different ETBs. } 
	\label{fig:fourfigures}
	\vspace{-10pt}
\end{figure}

\noindent{\textbf{Effect of Stage Number.}}
We analyze the influence of the number of search stages in our method. We test our GBi-Net model on DTU 
dataset with a maximum stage
number of 9. We compare the reconstruction results of Stage 6, 7, 8, 9 with both the point cloud evaluation metrics and the depth map evaluation metrics.
As shown in Fig. ~\ref{fig:Depthmapevaluationsub} 
and Fig. ~\ref{fig:Pointcloudevaluationsub},
the reconstruction results improve quickly 
from Stage 6 to 8 and then convergence, which indicates that our model can converge with a reasonably small stage number.

% \begin{table}[htbp]
%   \centering
%   \caption{Point cloud evalution and depth map evaluation of results of different stage numbers on DTU.}
%     \resizebox{\linewidth}{!}{\begin{tabular}{cccccccc}
%     \toprule
%     Stage No. & Acc. $\downarrow$& Comp. $\downarrow$& Overall $\downarrow$& $<$0.125$\uparrow$ & $<$0.25$\uparrow$ & $<$0.5$\uparrow$ & $<$1$\uparrow$ \\
%     \midrule
%     6     & 0.512 & 0.263 & 0.388 & 4.859 & 9.714 & 19.38 & 38.38 \\
%     7     & 0.368 & 0.242 & 0.305 & 8.615 & 17.16 & 33.63 & 59.88 \\
%     8     & 0.315 & 0.262 & 0.289 & 12.77 & 24.89 & 45.10  & 65.94 \\
%     9     & 0.316 & 0.265 & 0.291 & 13.34 & 25.86 & 45.91 & 65.47 \\

%     \bottomrule
%     \end{tabular}}%
%   \label{tab:iterationnumber}%
% \end{table}%

% \begin{figure}[t]
%   \centering
%   \includegraphics[width=0.8\linewidth]{image/mask_bin_plot.pdf}
% \vspace{-10pt}
%   \caption{Valid pixel percentage and $<1mm$ percentages of models with different Depth numbers in different stages. The valid pixel percentage is slightly below 100\% in Stage 1 because there are several out of range depths in the ground-truth data.}
%   \label{fig:mask_bin_plot}
% \end{figure}

\begin{table}[htbp]
  \centering
  \caption{Evaluation on the number of Error Tolerance Bin (ETB), considering the performance of point cloud, depth map, and the memory usage on DTU. 0, 1 and 2 indicate respectively adding 0, 1, 2 ETBs on both sides of the search bins.}
  \vspace{-5pt}
    \resizebox{\linewidth}{!}{\begin{tabular}{ccccccccc}
    \toprule
    \# of ETBs. & Overall $\downarrow$& $<$0.125$\uparrow$ & $<$0.25$\uparrow$ & $<$0.5$\uparrow$ & $<$1$\uparrow$ & Mem.(MB) \\
    \midrule
    0   & 0.360    & 5.622 & 11.19 & 21.91 & 39.14 & \textbf{2108} \\
    1    & \textbf{0.298} & 11.75 & 22.92 & 41.62 & 60.99 & \textbf{2108} \\
    2    & 0.300   & \textbf{13.58} & \textbf{26.18} & \textbf{45.78} & \textbf{63.92} & 2438 \\
    \bottomrule
    \end{tabular}}%
  \label{tab:differentclassnumbers}%
%   \vspace{-8pt}
\end{table}%

\begin{table}[htbp] 
  \centering
  \caption{Point cloud evaluation and training memory overhead comparison on DTU 
  \cite{aanaes2016large} for different gradient updating strategies. GBi-Net w/GC and GBi-Net w/EGU indicate the proposed GBi-Net with commonly used gradient accumulation and the proposed efficient gradient updating, respectively.}

\vspace{-6pt}
\resizebox{0.93\linewidth}{!}{
    \begin{tabular}{ccccc}
    \toprule
    Method & Acc. & Comp. & Overall & Memory (MB) \\
    \midrule
    GBi-Net w/ GC & \textbf{0.320}  & 0.277 & 0.299 & 12137 \\
    GBi-Net w/ EGU & {0.326} & \textbf{0.269} & \textbf{0.298} & \textbf{5208} \\
    \bottomrule
    \end{tabular}
}%
  \label{tab:memoryefficienttraining}%
  \vspace{-8pt}
\end{table}%

\noindent{\textbf{Effect of ETB Number.}}
We evaluate our generalized binary search with
different numbers of error tolerance bins on DTU. 
% As described in Sec.~\ref{sec:generalizedbinarysearch},
% different ETB numbers correspond to different depth hypotheses,\ie~$D$. 
We train 3 networks with 0, 1 and 2 ETBs on both sides, \ie~2, 4 and 6 depth hypotheses respectively. All of the experiments run with the same setting. 
% The 0 ETB (\ie~$D=2$) model is our proposed binary search, and 1 ETB (\ie~$D=4$) model is the generalized binary search proposed in this paper.
% The 2 ETB (\ie~$D=6$) model is to add 2 ETBs on both sides.
The evaluation results and the memory consumption are shown in Table~\ref{tab:differentclassnumbers}.
Fig.~\ref{fig:Validpixelpercentagesub} and Fig.~\ref{fig:1mmpercentagesub} also shows the valid pixel percentage and $<$1mm pixel percentage of different models in different stages.
% The result of $D=4$ is much better than those of $D=2$.
From 0 ETB to 1 ETB, we can observe a significant improvement in accuracy. This reveals the importance of ETBs.
Note that the memory consumption of 
1 ETB remains the same level as 0 ETB because the GPU memory is still dominated by the 2D image encoder.
In Table~\ref{tab:differentclassnumbers}, 
2 ETBs model is slightly better than 1 ETB model on the depth map metric while consuming more memory. It also does not show better performance on the point cloud metric. Besides, in Fig.~\ref{fig:1mmpercentagesub}, the valid and $<$1mm pixel percentages of 1 ETB and 2 ETBs models are very close. All these results reveal that 1 ETB in our GBi-Net is already sufficient to achieve a good balance between accuracy and memory usage. More importantly, further increasing ETBs may make the model more complex and harder to optimize. 
% Also we can see that in depth $D=4$ model, there are still $69.2\%$
% valid pixels after the final stage. 

\noindent{\textbf{Effect of memory-efficient Training.}}
We train a model without our Memory-efficient Training
strategy (see Sec.~\ref{sec:networkoptimization}) on the DTU dataset.~This model averages the gradients accumulated from all the different stages and back-propagates them, which is a widely performed gradient updating scheme in existing MVS methods. As shown in Table~\ref{tab:memoryefficienttraining}, these two comparison models obtain very similar results on the depth performance, while with the proposed Memory-efficient Training, the memory consumption is largely reduced by $57.1\%$.

\section{Conclusion}
In this paper, we first presented a binary search network (Bi-Net) design for MVS to significantly reduce the memory footprint of 3D cost volumes. Based on this design, we further proposed a generalized binary search network (GBi-Net) containing three effective mechanisms, \ie~error tolerance bins, gradients masking, and efficient gradient updating.
The GBi-Net can greatly improve the accuracy while maintaining the same
memory usage as the Bi-Net.
Experiments on challenging datasets also showed state-of-the-art depth prediction accuracy, and remarkable memory efficiency of the proposed methods.

% \noindent{\textbf{Loss Function.}}
% We also train a model with regression loss commonly used in learning-based
% MVS \cite{yao2018mvsnet, gu2020cascade}.
% Since the regressed depths could be arbitrary values, 
% we update the depth hypotheses with a slightly different
% strategy. After each stage, we the narrow the depth range
% by half, with the current depth value as center.
% Then we sample 4 depth hypotheses in this range.
% We evaluate this models on DTU dataset.
% Figure 
% \noindent{\textbf{Loss Function.}}
% We also train a model with regression loss commonly used in learning-based
% MVS \cite{yao2018mvsnet, gu2020cascade}.
% Since the regressed depths could be arbitrary values, 
% we update the depth hypotheses with a slightly different
% strategy. After each stage, we the narrow the depth range
% by half, with the current depth value as center.
% Then we sample 4 depth hypotheses in this range.
% We evaluate this models on DTU dataset.
% As shown in Table \ref{tab:memoryefficienttraining},
% the classification loss outperforms regression loss.
% Another problem of model trained by regression loss 
% is that there are only 4 depths in each stage 
% so the photometric consistency defined in \cite{yao2018mvsnet}
% is no longer valid.

\newpage
\clearpage
\begin{center}
      {\Large \bf Supplementary material}
\end{center}
\setcounter{section}{0}

In this supplementary, we introduce details about the depth map fusion procedure and provide more qualitative results regarding the ablation study and the overall performance of the proposed model. 
\section{Depth map Fusion}
\label{sec:depthmapfusion}
As indicated in 
% Sec. 4.2 of 
the main paper,
after obtaining the final depth maps of a scene, we filter and fuse depth maps into one point cloud.
The final depth maps are generated 
from center points of the selected
bins of the final stage.
% Similar to previous MVS methods \cite{yao2018mvsnet, yao2019recurrent},
We consider both the photometric and the geometric
consistency for depth map filtering.
The geometric consistency is similar to MVSNet~\cite{yao2018mvsnet} measuring the depth consistency among multiple views. The photometric consistency, however, is different.
The probability volume $\textbf{P}$
is considered to construct the photometric consistency, following R-MVSNet~\cite{yao2019recurrent}. As the probability volume
is the classification probabilities for the depth hypotheses, it measures the matching quality of these hypotheses.
Since the proposed method consists of $K$ stages, we can obtain $K$ probability volumes, \ie~$\{\textbf{P}_{k}| k=1,...,K\}$.
For each pixel $\textbf{p}$, its photometric consistency
from its $K$ probabilities can be calculated as follows:
\begin{equation}\label{equ:photometricconsistency}
    Ph(\textbf{p}) = \frac{1}{K'}\sum_{k=1}^{K'}\max \{\textbf{P}_{k}(j, \textbf{p}) | j=1,...,D\}.
\end{equation}
Where $Ph(\textbf{p})$ is the photometric consistency
of pixel $\textbf{p}$; $D$ is depth hypothesis number; 
The $\max$ operation obtains the classification probability of
a selected hypothesis; $K'$ is the maximum stage considered in 
photometric consistency and $1 \leq K' \leq K$.
Equation~\ref{equ:photometricconsistency}
actually computes an average of 
the probabilities of the $K'$ stages.
In practice, when the maximum stage number
$K=8$, we set $K'=6$. It means that we take the average probability of the first 6 stages as the score of the photometric consistency.
In our multi-stage search pipeline,
as the resolutions of probability volumes are different, we upsample them to the maximum resolution of stage $K$ before the computation.
After producing the photometric consistency score for each pixel, the depths of pixels are discarded if their consistency scores are below a threshold.

\begin{figure*}[t]
  \centering
   \includegraphics[width=\linewidth]{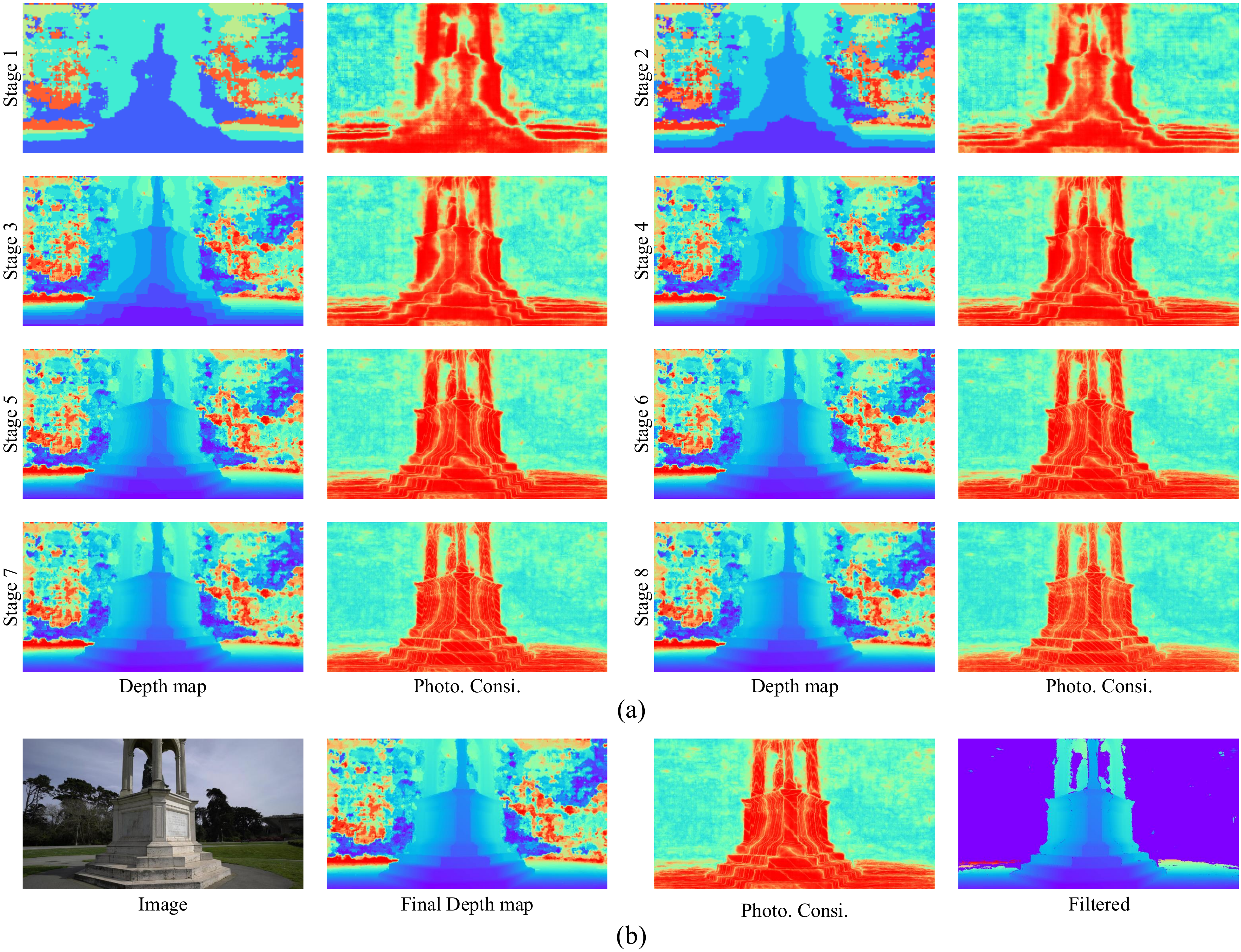}
   \caption{
(a) The predicted depth maps and
photometric consistency (Photo. Consi.) maps
in all the stages of a sample in Tanks and Temples \cite{knapitsch2017tanks}.
(b) The input image, final predicted depth map of Stage 8, 
photometric consistency (Photo. Consi.) map of Stage 6
and filtered depth map by photometric consistency.
The background of this image is far away from the 
foreground and is out of the depth range so MVS
methods will predicted outlier values for background pixels.
With the photometric consistency, we can effectively filter out
these outliers.
% The background of this image is far away from the 
% foreground and is out of the depth range so MVS
% methods will predicted outlier values for such pixels.
% With the photometric consistency, we can effectively filter out
% these outliers.
}
   \label{fig:depthmap_tanks}
\end{figure*}

Figure~\ref{fig:supp_dtu_depthmaps}a in the supplementary shows the results of
each stage of a sample in the DTU dataset~\cite{aanaes2016large}.
The depth map in each stage consists of the center-point depth values
of selected bins.
The quality of these depth maps can be improved quickly, demonstrating a fast search convergence of our method.
The valid mask maps represent valid pixels in each search stage. Note that these mask maps are combined with the ground-truth mask maps from the dataset, and thus the background
pixels are not considered.
The photometric consistency (Photo. Consi.) map
in stage $k$ is computed using Equation~\ref{equ:photometricconsistency}
by setting $K'=k$.
As shown in the Figure~\ref{fig:supp_dtu_depthmaps}a, the photometric consistency maps
is an effective measurement of depth map quality.
As shown in Figure~\ref{fig:supp_dtu_depthmaps}b, the photometric consistency (Photo. Consi.) maps
from Stage 6 are used to filter the final depth maps produced from Stage 8. The filtered depth maps are further refined by geometric consistency map, and finally fused into one point cloud.
Figure~\ref{fig:depthmap_tanks} also shows qualitative results of a sample from the Tanks and Temples~\cite{knapitsch2017tanks} dataset.
The background of this image is far away from the 
foreground and is out of the depth range, so the MVS
methods predict outlier values for the background pixels. Using the photometric consistency maps, we can effectively filter out
these outliers.

% \section{Evaluation on Video Dataset}

\section{More Visualization Results}
\begin{figure*}[t]
  \centering
   \includegraphics[width=\linewidth]{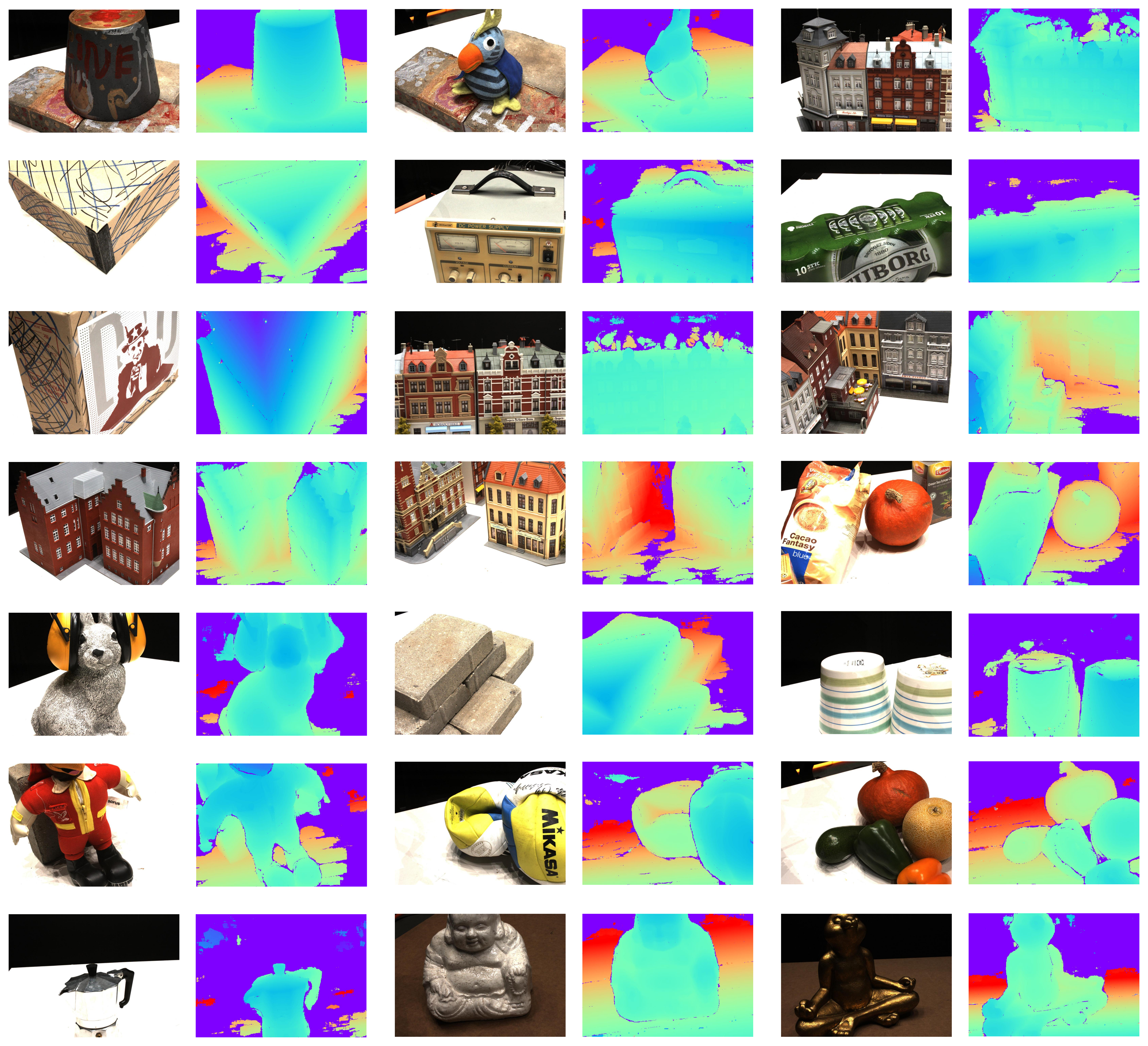}
   \caption{Examples of images and their corresponding depth maps in DTU dataset \cite{aanaes2016large}. The depth maps are filtered by photometric consistency.
}
   \label{fig:depthmap_final_dtu}
\end{figure*}

\begin{figure*}[t]
  \centering
   \includegraphics[width=\linewidth]{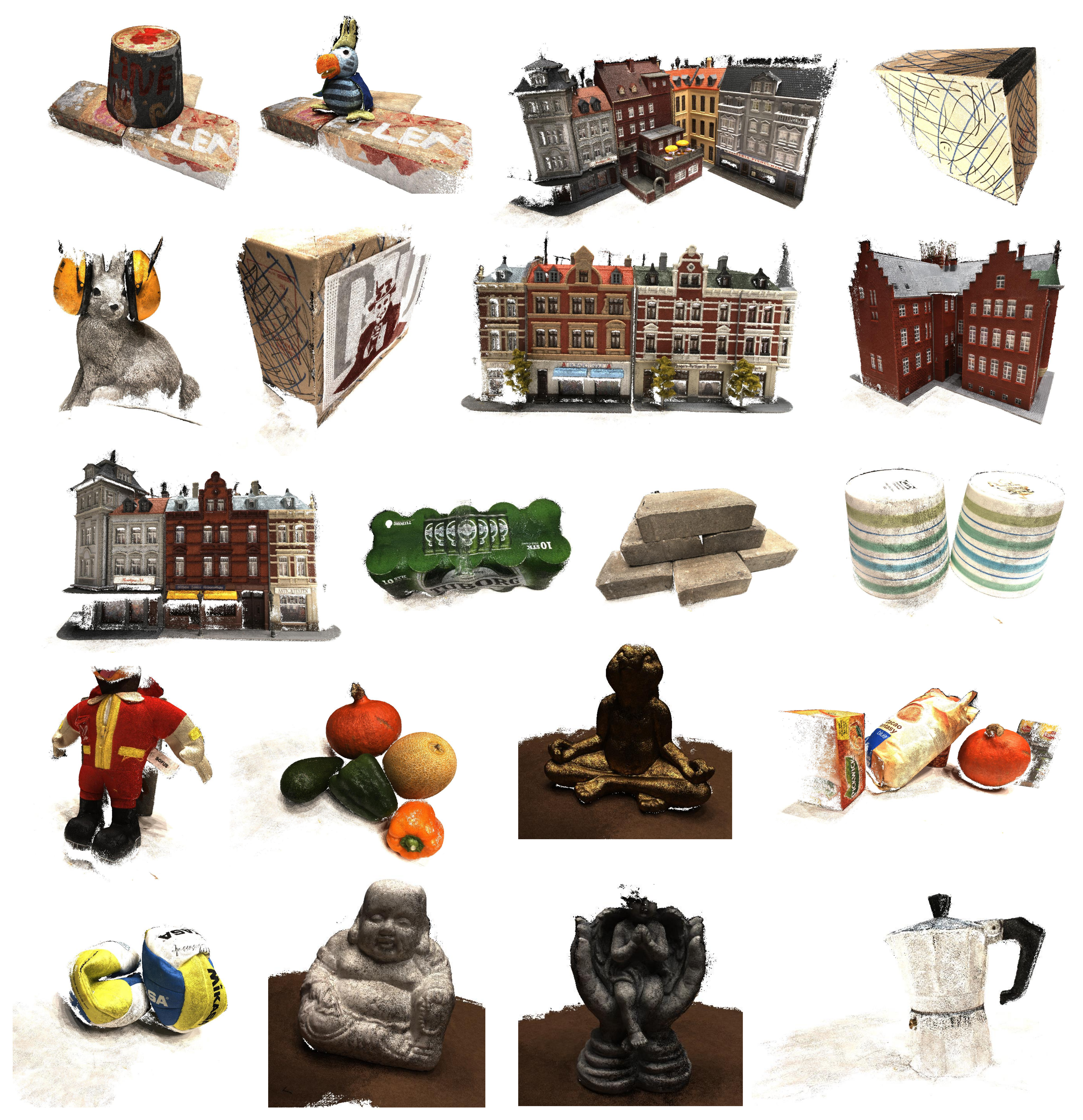}
   \caption{
   Point clouds of our method on DTU dataset  \cite{aanaes2016large}.
%   The detailed updating step can be seen in Fig.\ref{fig:quadraticsearch}.
}
   \label{fig:supp_dtu_pt}
\end{figure*}

\begin{figure*}[t]
  \centering
   \includegraphics[width=\linewidth]{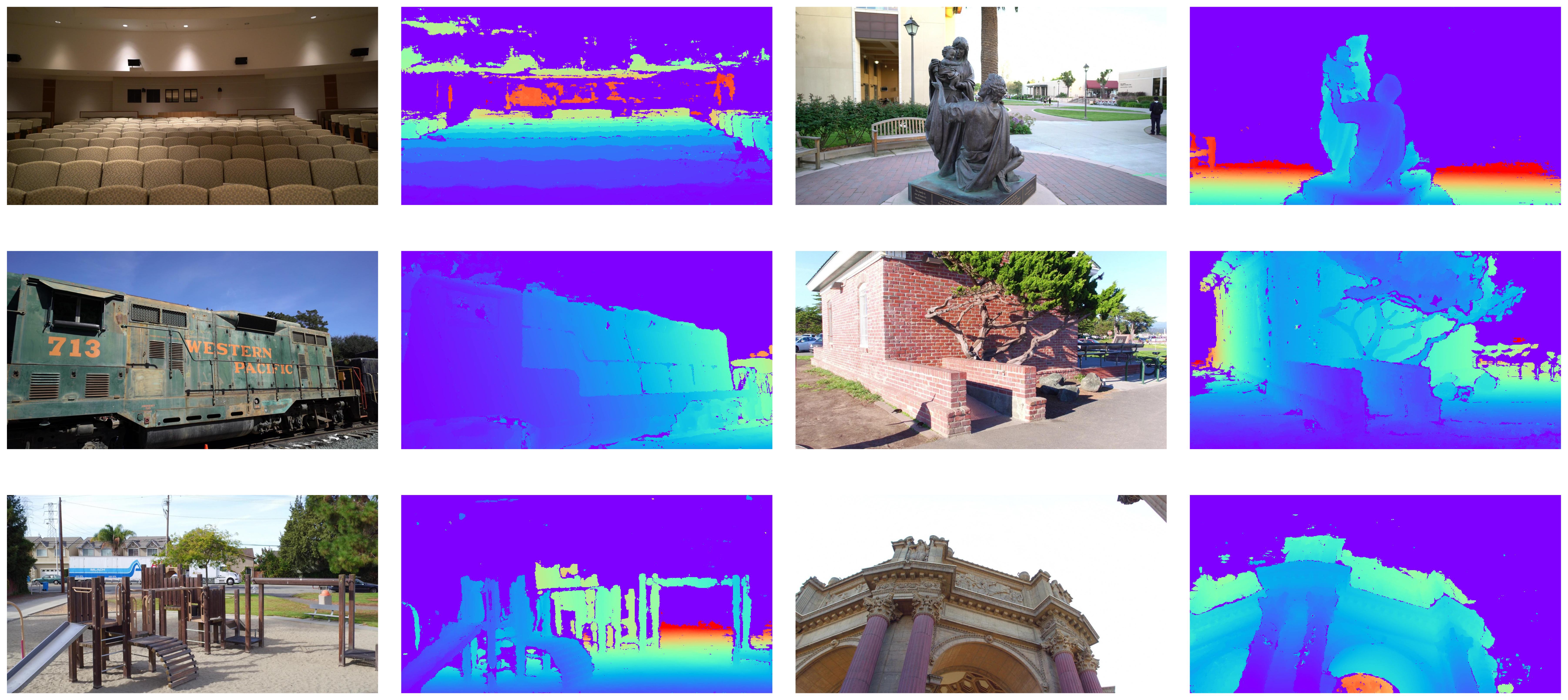}
   \caption{
   Examples of images and their corresponding depth maps in Tanks and Temples dataset \cite{knapitsch2017tanks}. The depth maps are filtered by photometric consistency.
%   The detailed updating step can be seen in Fig.\ref{fig:quadraticsearch}.
}
   \label{fig:depthmap_final_tanks}
\end{figure*}

\begin{figure*}[t]
  \centering
   \includegraphics[width=\linewidth]{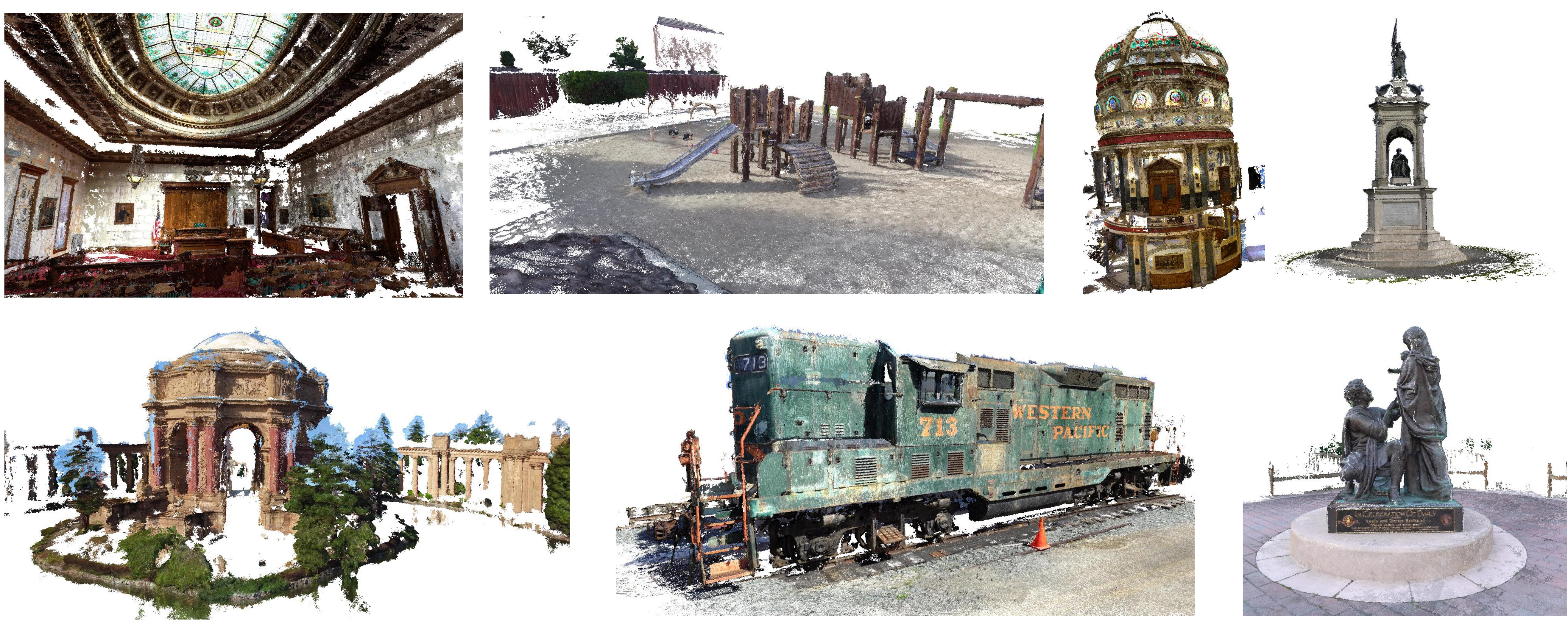}
   \caption{
   Point clouds of our method on Tanks and Temples dataset \cite{knapitsch2017tanks}.
%   The detailed updating step can be seen in Fig.\ref{fig:quadraticsearch}.
}
   \label{fig:supp_tanks_pt}
\end{figure*}

We show more qualitative results of the proposed model in this section. Figure~\ref{fig:depthmap_final_dtu} and Figure~\ref{fig:depthmap_final_tanks} show several images and their corresponding depth maps in DTU dataset \cite{aanaes2016large} and Tanks and Temples dataset \cite{knapitsch2017tanks} respectively. The depth maps are filtered by photometric consistency. Figure~\ref{fig:supp_dtu_pt} and Figure \ref{fig:supp_tanks_pt} shows several point clouds of our method in DTU dataset \cite{aanaes2016large} and Tanks and Temples dataset \cite{knapitsch2017tanks} respectively.

%  Figure \ref{fig:supp_tanks_pt} shows several point clouds of our method on Tanks and Temples dataset \cite{knapitsch2017tanks}. 

% Fig.~\ref{fig:depthmapexample} shows a visualization result of our
% method.
% \subsection{Photometric Consistency}
% Figure of classification probability for each stage.
% Figure of final consistency for each stage.

% \clearpage

\begin{figure*}[htbp]
  \centering
   \includegraphics[width=\linewidth]{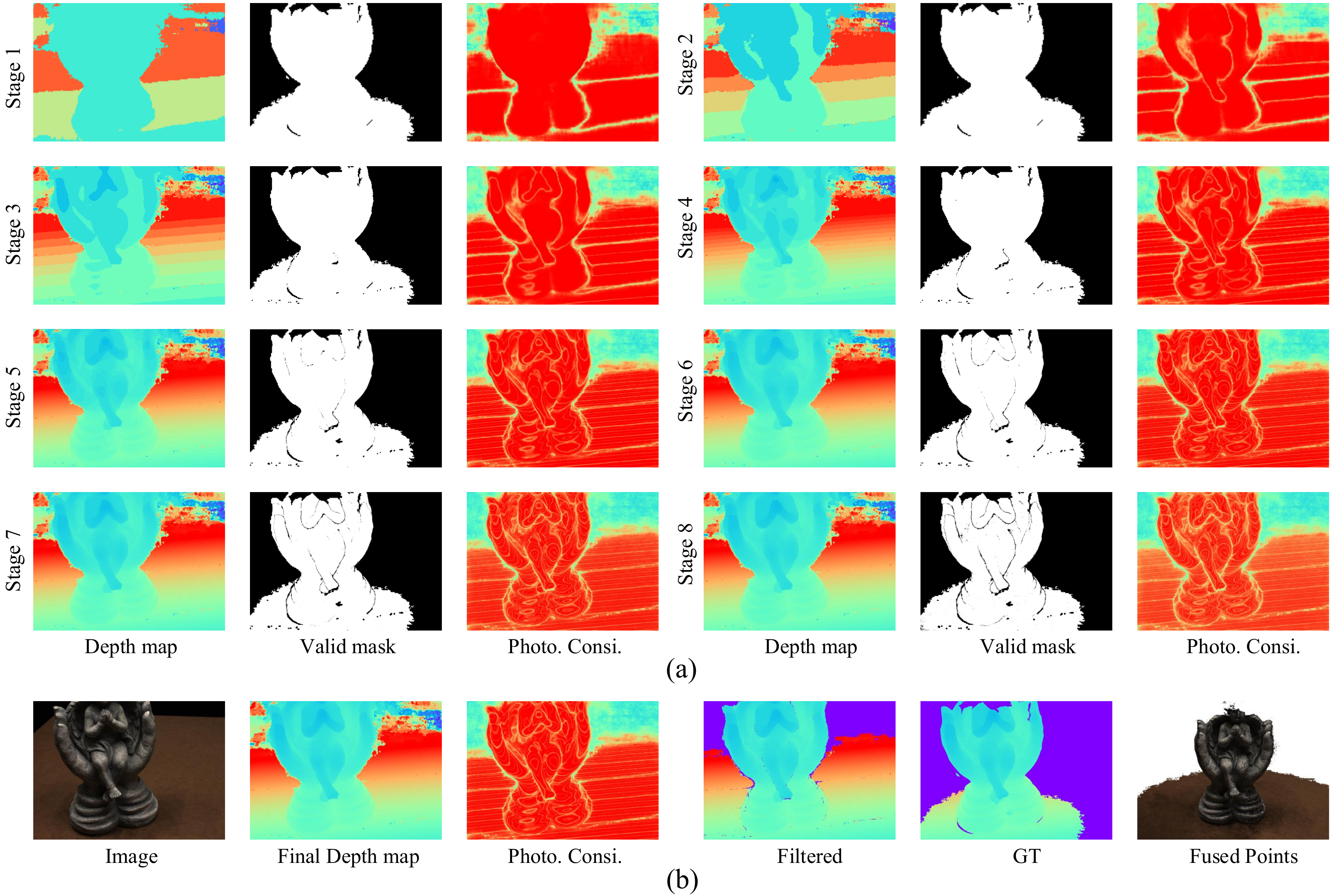}
   \caption{
(a) The predicted depth maps, valid mask maps and
photometric consistency (Photo. Consi.) maps
in all the stages of a sample in DTU \cite{aanaes2016large}.
(b) The input image, final predicted depth map of Stage 8, 
photometric consistency (Photo. Consi.) map of Stage 6,
filtered depth map by photometric consistency, ground truth
depth map and fused point cloud.
}
   \label{fig:supp_dtu_depthmaps}
\end{figure*}
\clearpage

\clearpage

\clearpage

\clearpage
\clearpage

%%%%%%%%% REFERENCES
{\small
\bibliographystyle{ieee_fullname}
\bibliography{egbib}
}

\end{document}